\documentclass[10pt,twocolumn,letterpaper]{article}

\usepackage{cvpr}
\usepackage{times}
\usepackage{epsfig}
\usepackage{graphicx}
\usepackage{amsmath}
\usepackage{amssymb}
\usepackage{mathtools}
\usepackage{color}
\usepackage{booktabs}
\usepackage[caption=false]{subfig}

\usepackage{algorithm}
\usepackage{algpseudocode}
\algrenewcommand{\algorithmicrequire}{\textbf{Input:}}
\algrenewcommand{\algorithmicensure}{\textbf{Output:}}

\usepackage{mynotation}

\newcommand{\parsection}[1]{\noindent\textbf{#1:} }

\usepackage[pagebackref=true,breaklinks=true,letterpaper=true,colorlinks,bookmarks=false]{hyperref}

\cvprfinalcopy % *** Uncomment this line for the final submission

\def\cvprPaperID{4984} % *** Enter the CVPR Paper ID here

\ifcvprfinal\fi
\begin{document}

%%%%%%%%% TITLE
\title{ATOM: Accurate Tracking by Overlap Maximization}

\newcommand{\asep}{\hspace{6mm}}
\newcommand{\aand}{\hspace{6mm}}
\author{Martin Danelljan$^{*,1,2}$ \aand Goutam Bhat$^{*,1,2}$ \aand Fahad Shahbaz Khan$^{1, 3}$ \aand Michael Felsberg$^1$ \vspace{1.5mm}\\
	 \small $^1$ CVL, Link\"oping University, Sweden \asep $^2$ CVL, ETH Z\"urich, Switzerland \asep $^3$ Inception Institute of Artificial Intelligence, UAE
}

\maketitle
\thispagestyle{empty}

%%%%%%%%% ABSTRACT
\begin{abstract}
   While recent years have witnessed astonishing improvements in visual tracking robustness, the advancements in tracking accuracy have been limited. As the focus has been directed towards the development of powerful classifiers, the problem of accurate target state estimation has been largely overlooked. In fact, most trackers resort to a simple multi-scale search in order to estimate the target bounding box. We argue that this approach is fundamentally limited since target estimation is a complex task, requiring high-level knowledge about the object.  

We address this problem by proposing a novel tracking architecture, consisting of dedicated target estimation and classification components. High level knowledge is incorporated into the target estimation through extensive offline learning. Our target estimation component is trained to predict the overlap between the target object and an estimated bounding box. By carefully integrating target-specific information, our approach achieves previously unseen bounding box accuracy. We further introduce a classification component that is trained online to guarantee high discriminative power in the presence of distractors. Our final tracking framework sets a new state-of-the-art on five challenging benchmarks. On the new large-scale TrackingNet dataset, our tracker \textbf{ATOM} achieves a relative gain of $15\%$ over the previous best approach, while running at over $30$ FPS. Code and models are available at \url{https://github.com/visionml/pytracking}.
\end{abstract}

{\let\thefootnote\relax\footnote{{$^*$Both authors contributed equally.}}}

%%%%%%%%% BODY TEXT

\section{Introduction}

Generic online visual tracking is a hard and ill-posed problem. The tracking method must learn an appearance model of the target  online based on minimal supervision, often a single starting frame in the video. The model then needs to generalize to unseen aspects of the target appearance, including different poses, viewpoints, lightning conditions etc. The tracking problem can be decomposed into a \emph{classification} task and an \emph{estimation} task. In the former case, the aim is to robustly provide a coarse location of the target in the image by categorizing image regions into foreground and background. The second task is then to estimate the target state, often represented by a bounding box.

\begin{figure}[!t]
	\centering%
	\newcommand{\wid}{0.33\columnwidth}%
	\includegraphics*[trim = 0 0 0 0, width = \wid]{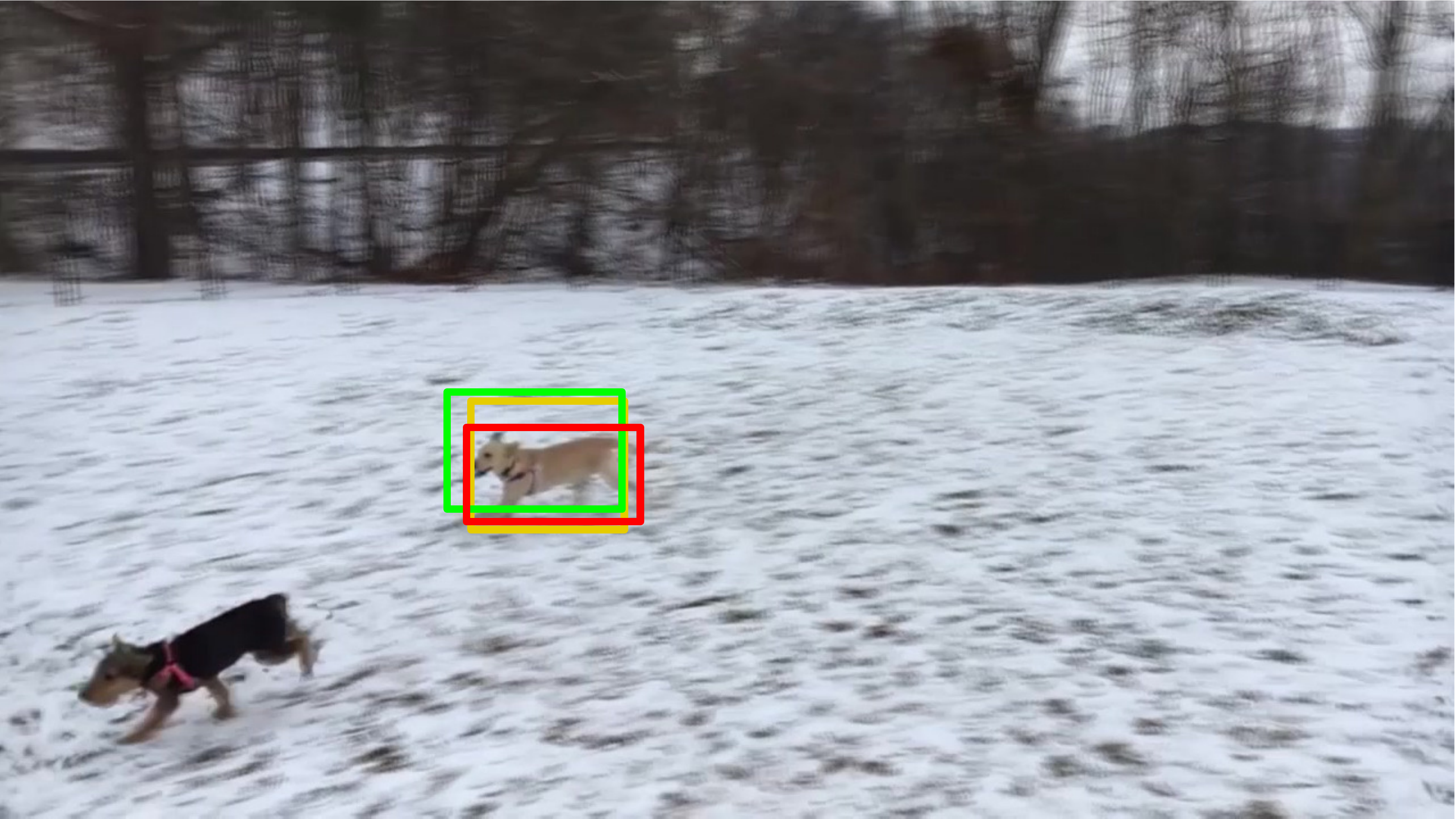}%
	\includegraphics*[trim = 0 0 0 0, width = \wid]{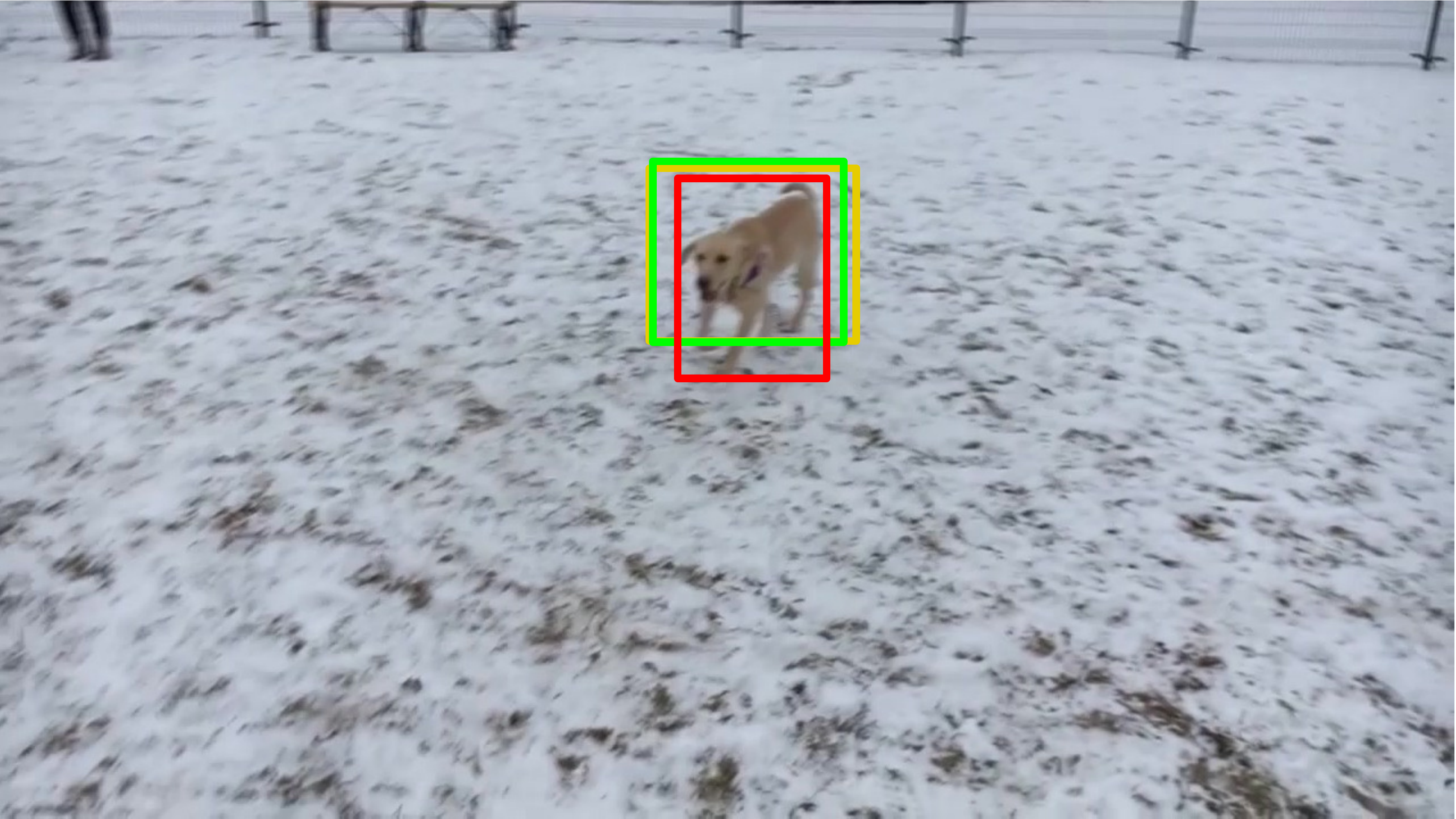}%
	\includegraphics*[trim = 0 0 0 0, width = \wid]{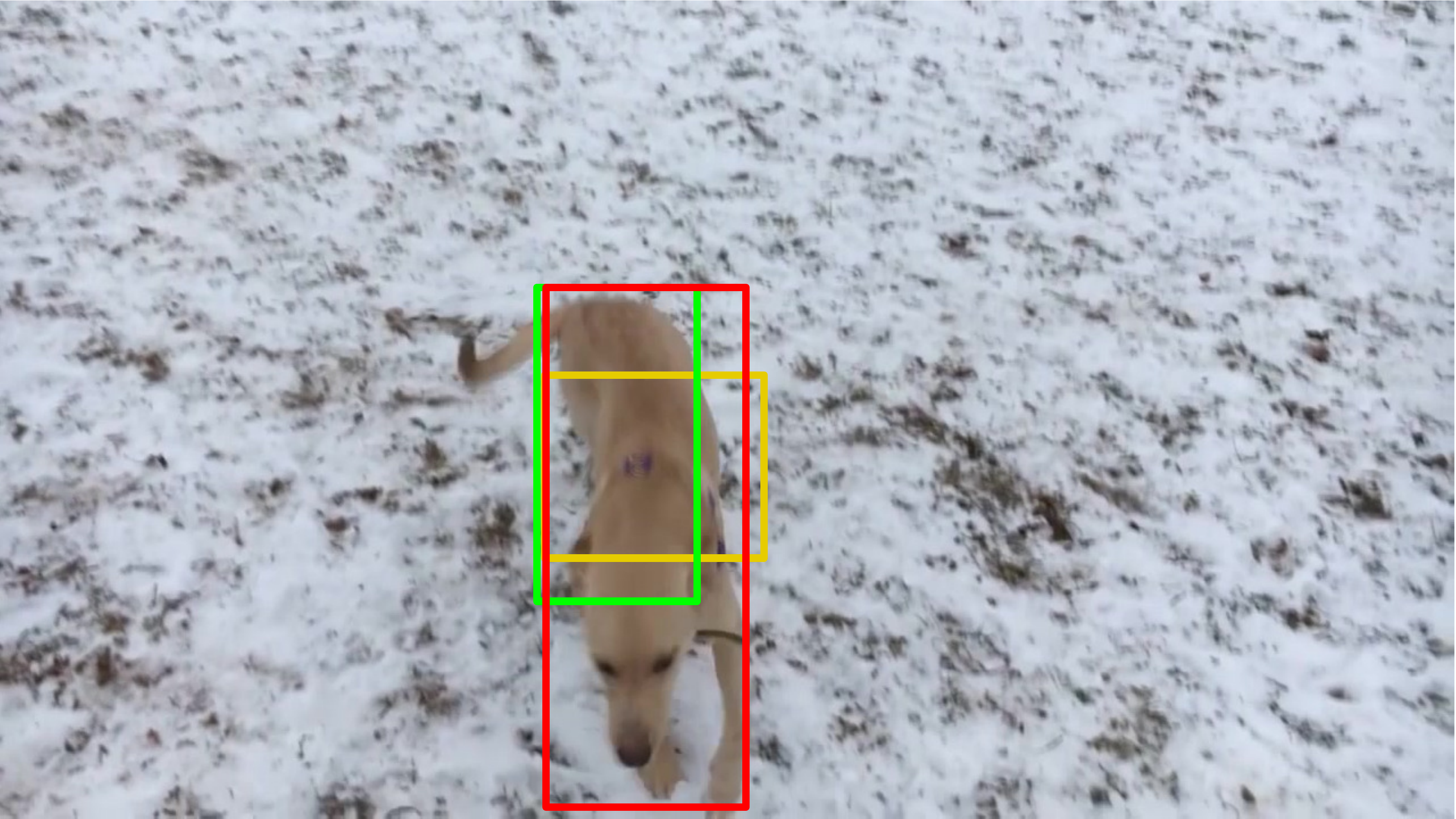}
	\includegraphics*[trim = 0 0 0 0, width = \wid]{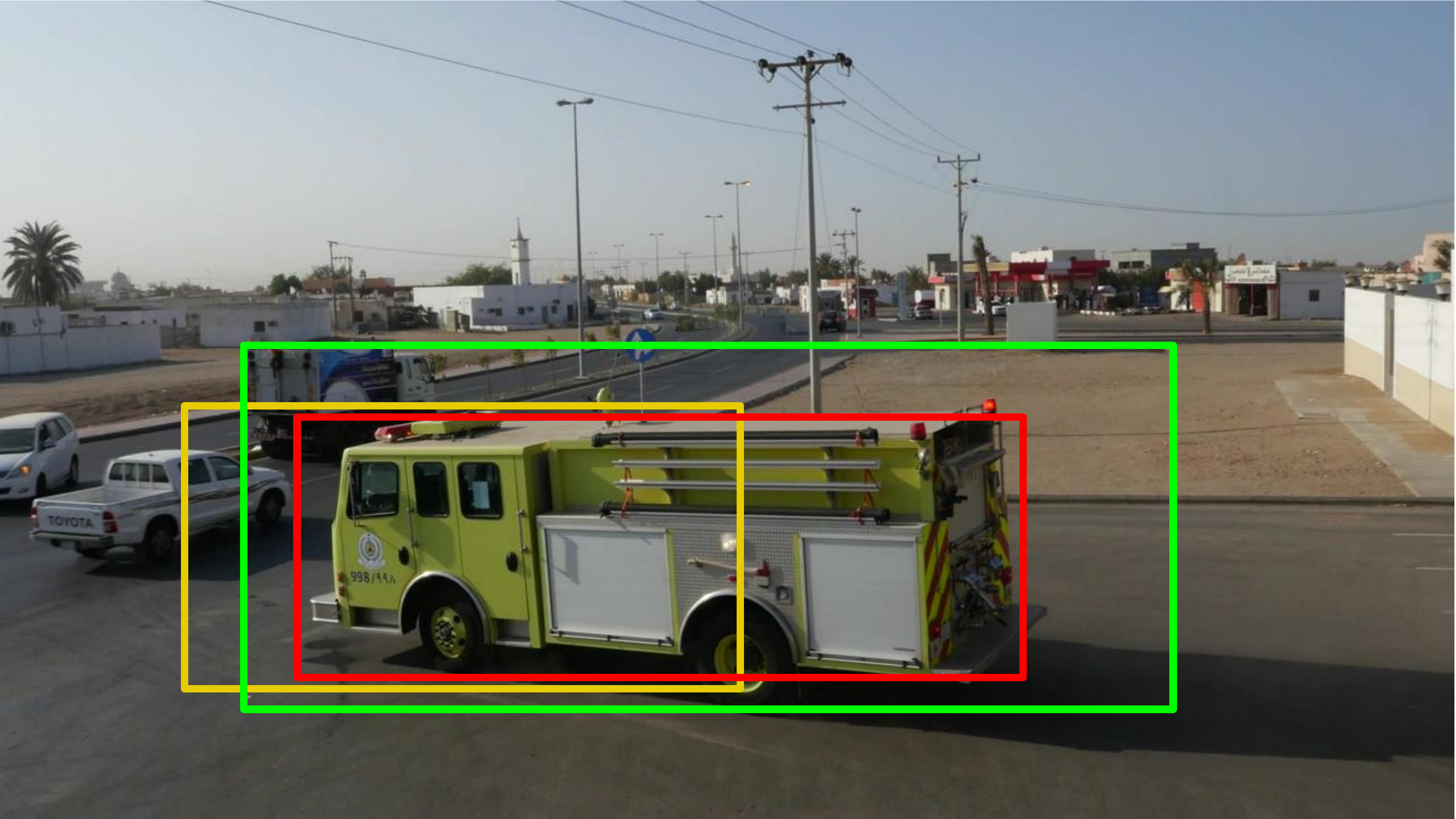}%
	\includegraphics*[trim = 0 0 0 0, width = \wid]{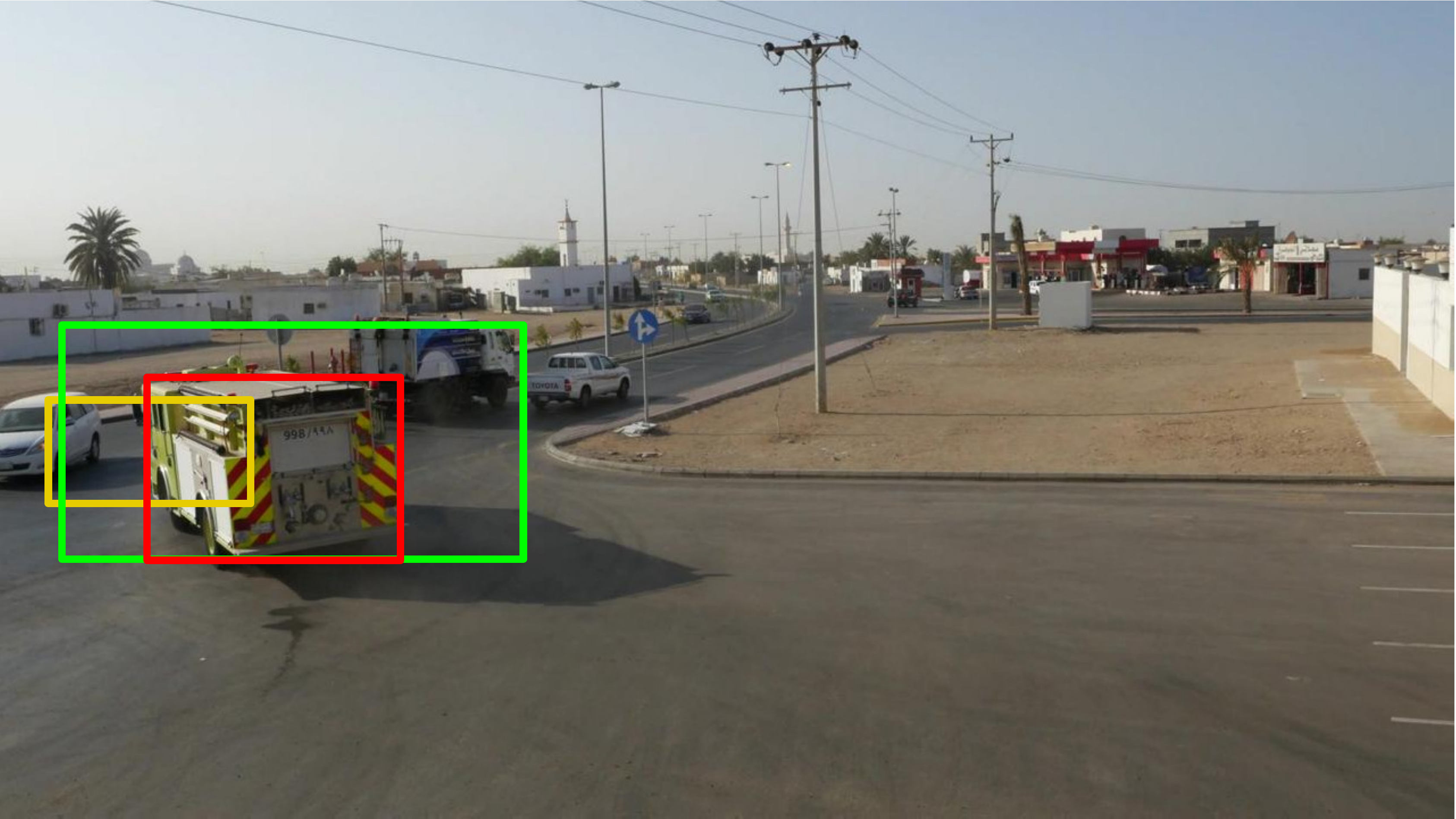}%
	\includegraphics*[trim = 0 0 0 0, width = \wid]{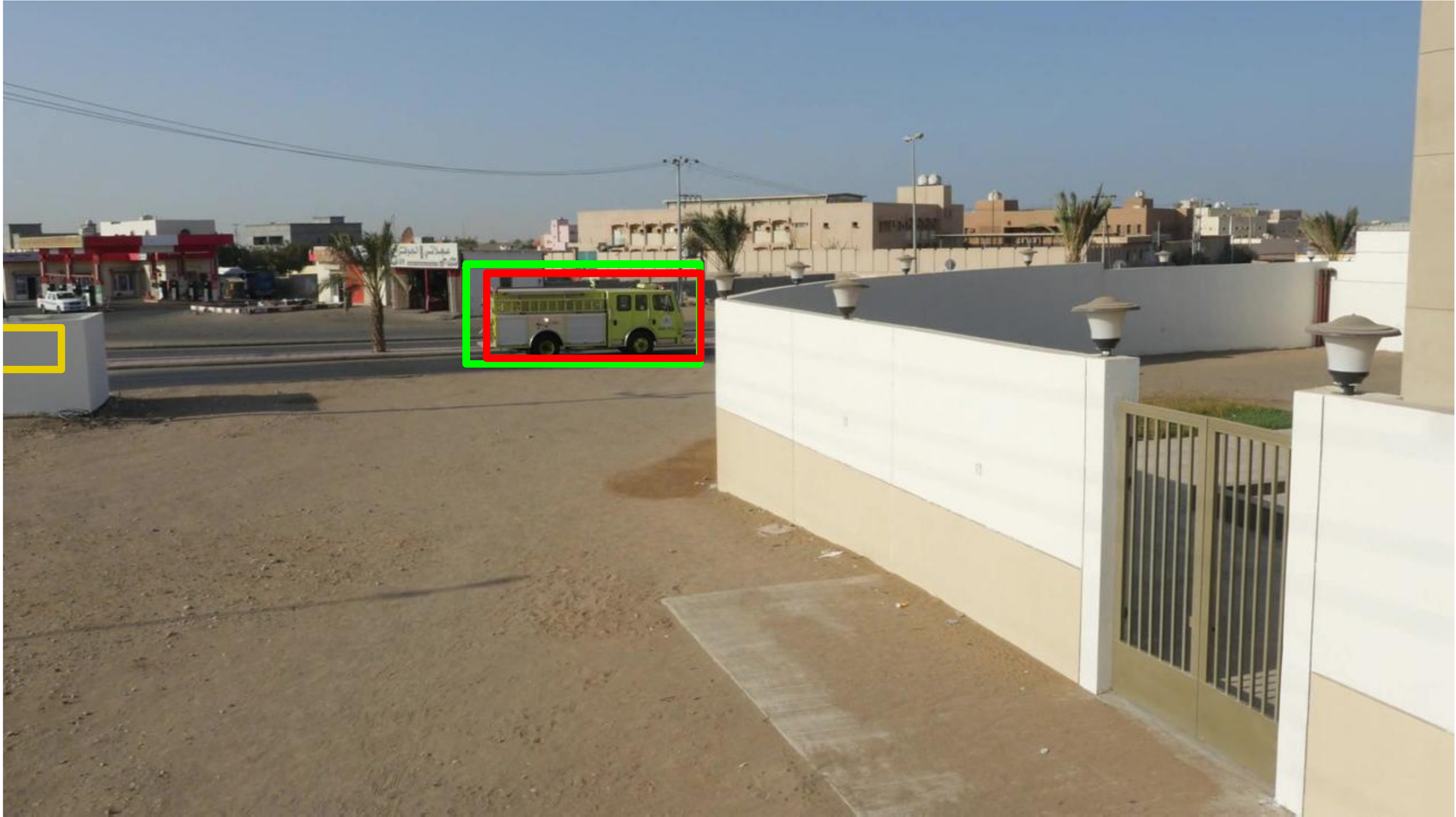}
	\includegraphics*[trim = 0 200 500 100, width = \wid]{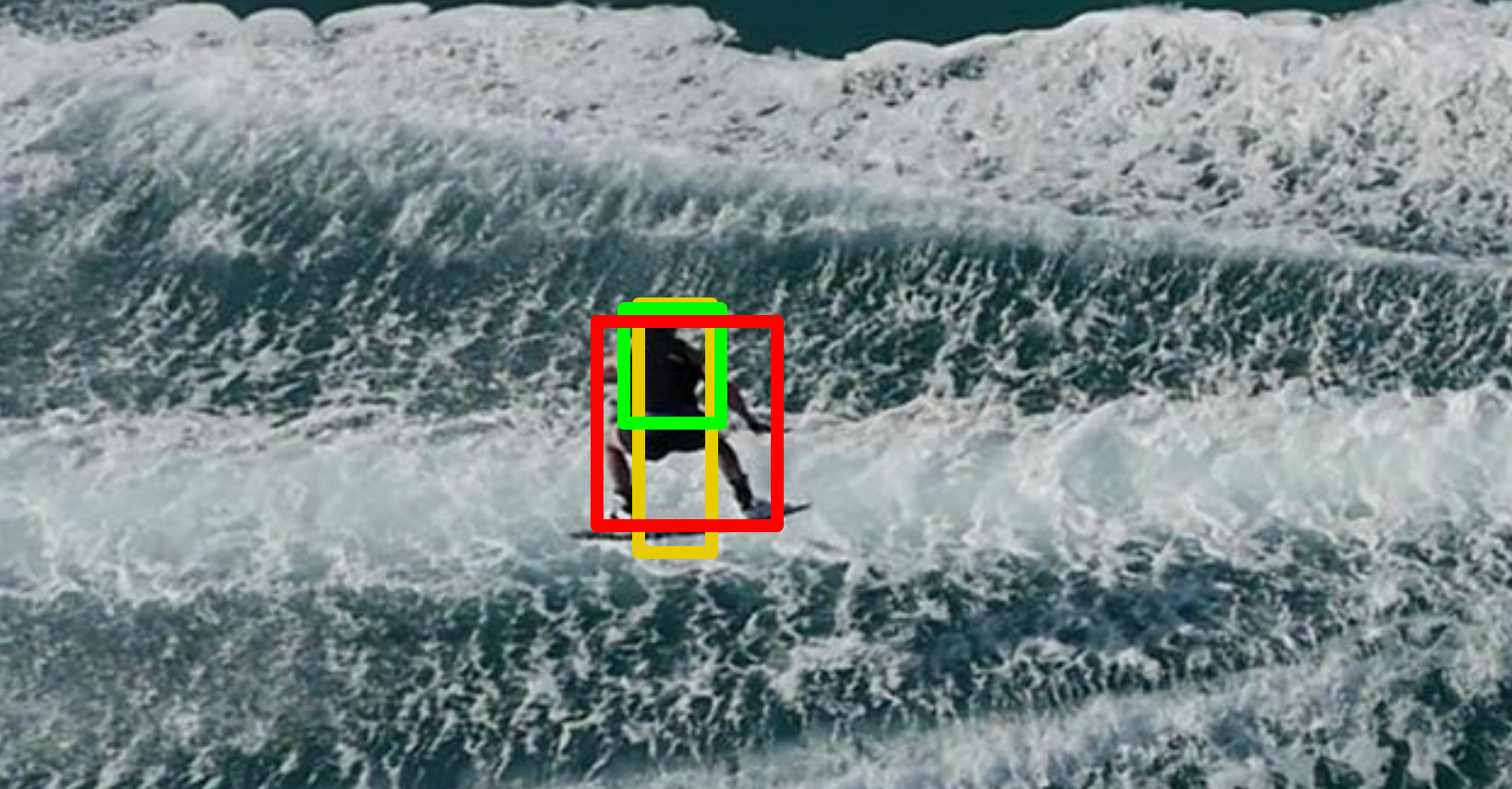}%
	\includegraphics*[trim = 100 200 400 100, width = \wid]{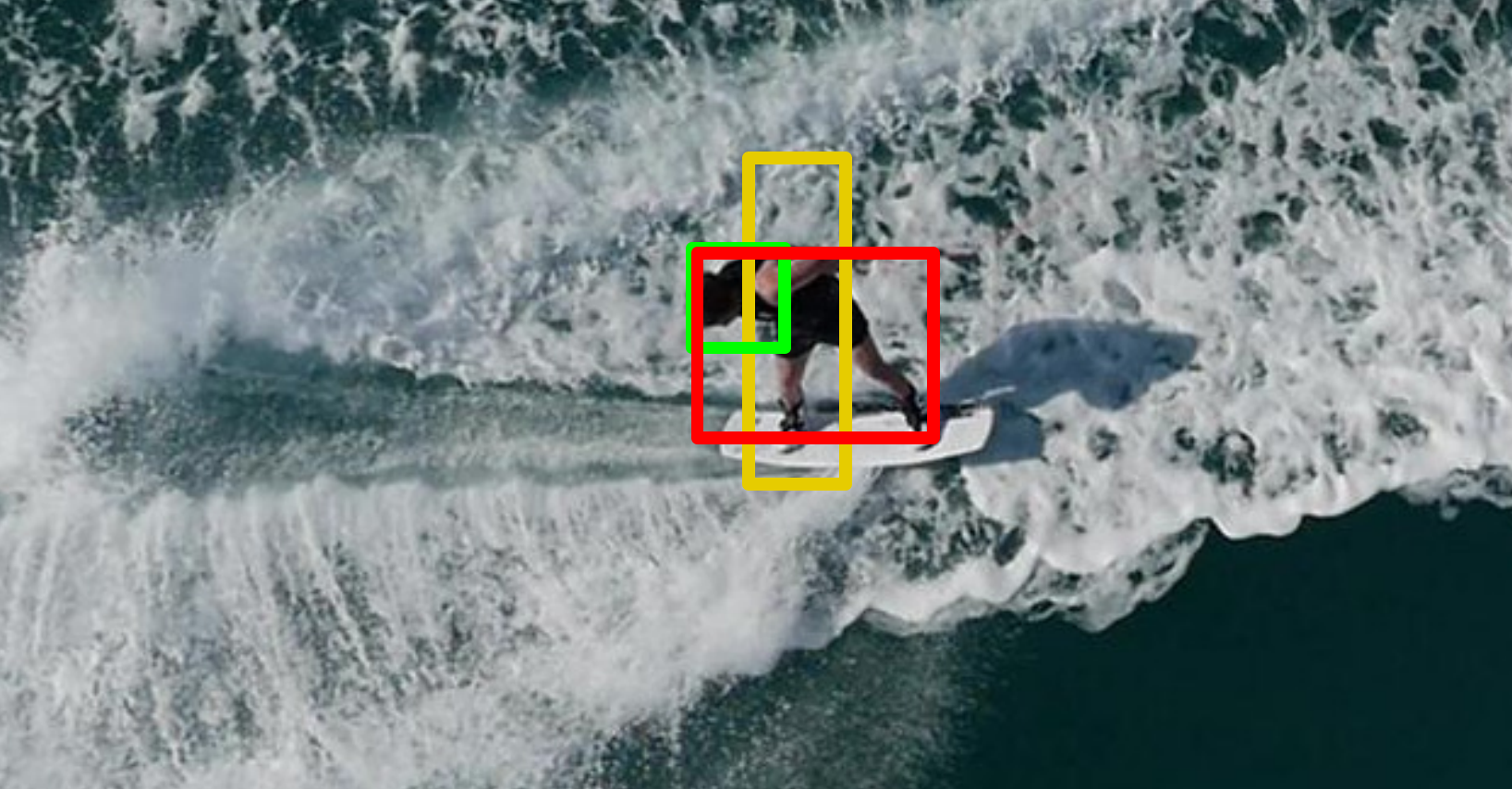}%
	\includegraphics*[trim = 400 300 100 0, width = \wid]{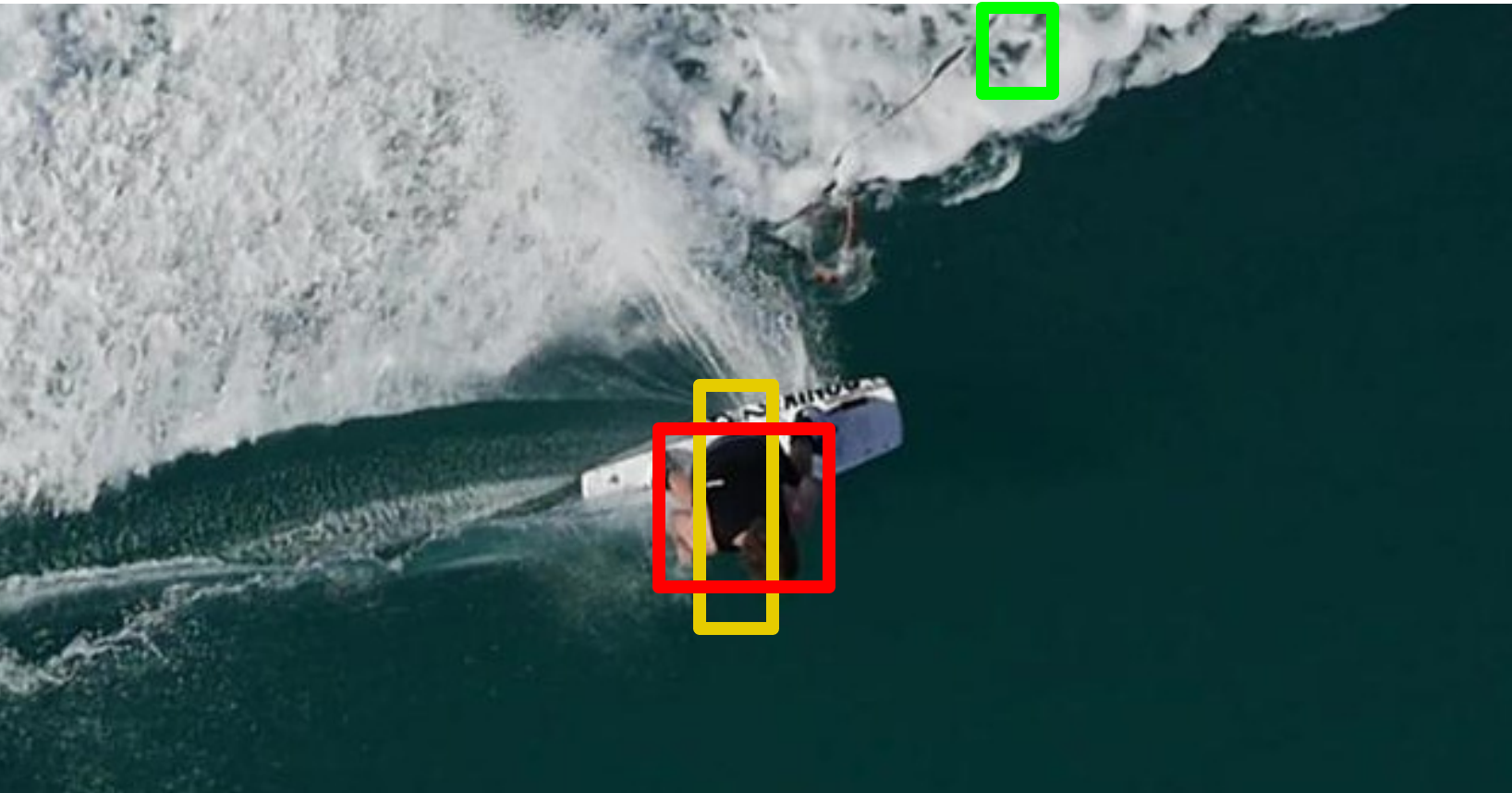}
	\includegraphics*[trim = 0 40 0 0, width = \wid]{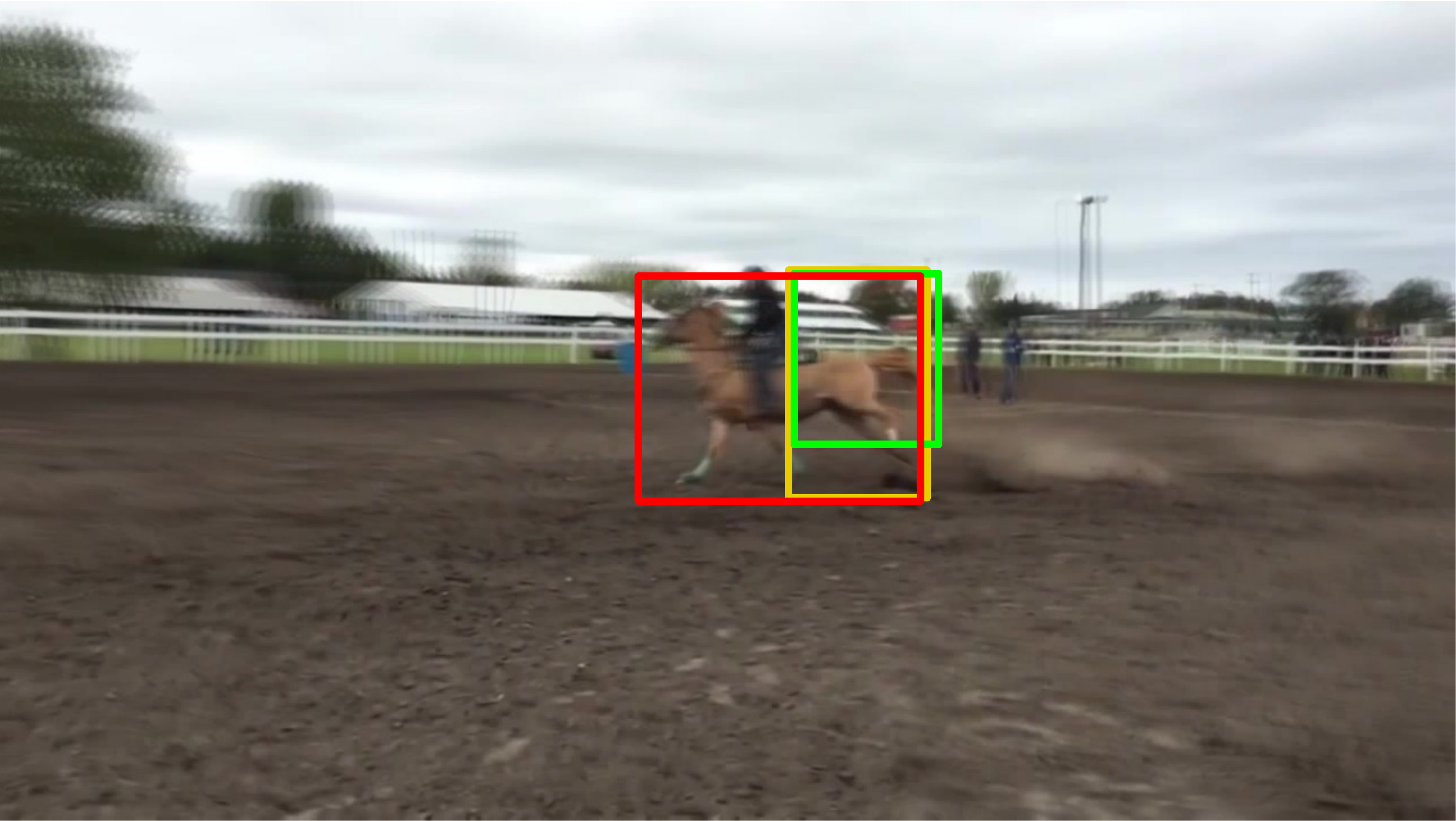}%
	\includegraphics*[trim = 0 40 0 0, width = \wid]{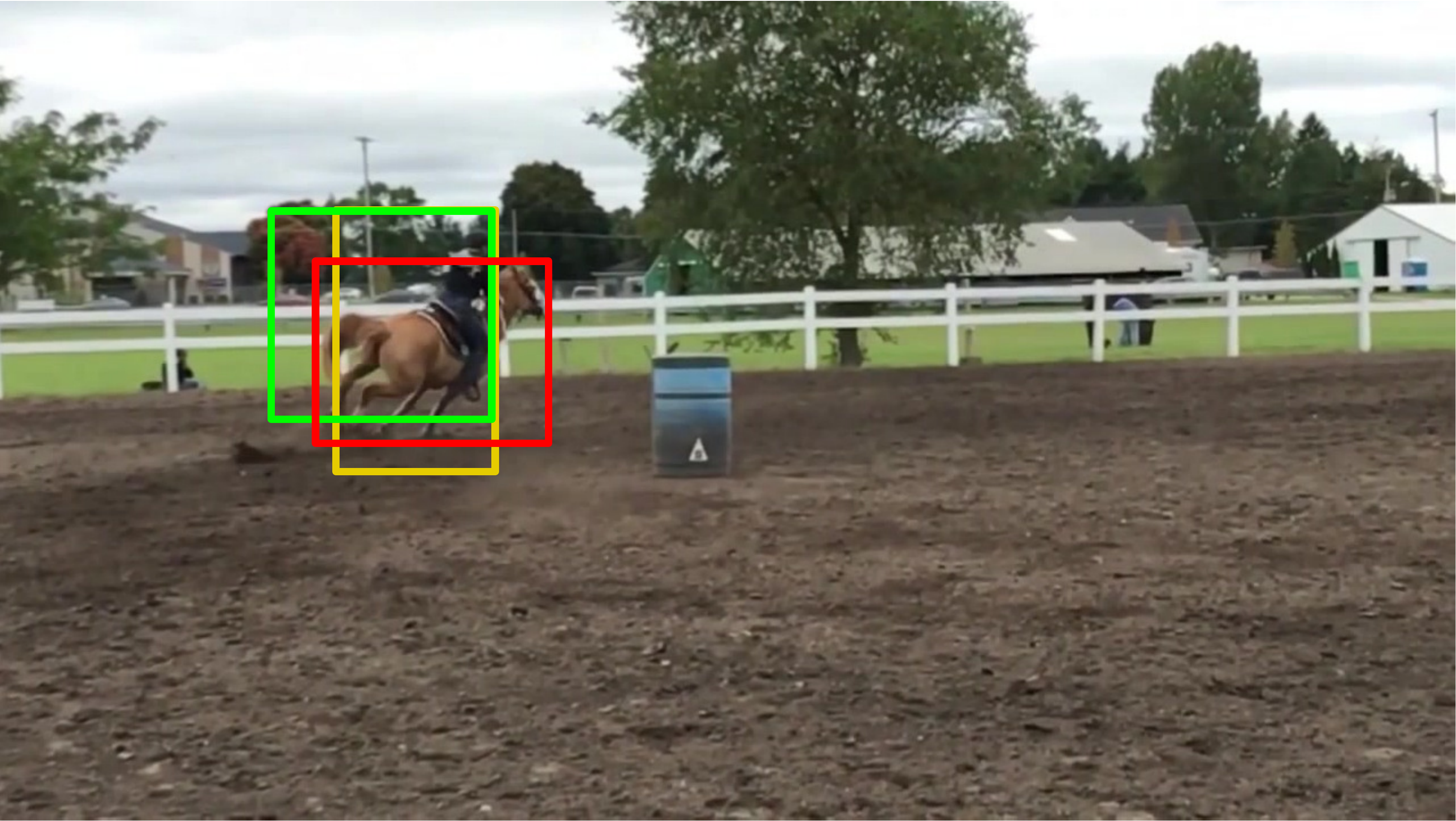}%
	\includegraphics*[trim = 0 40 0 0, width = \wid]{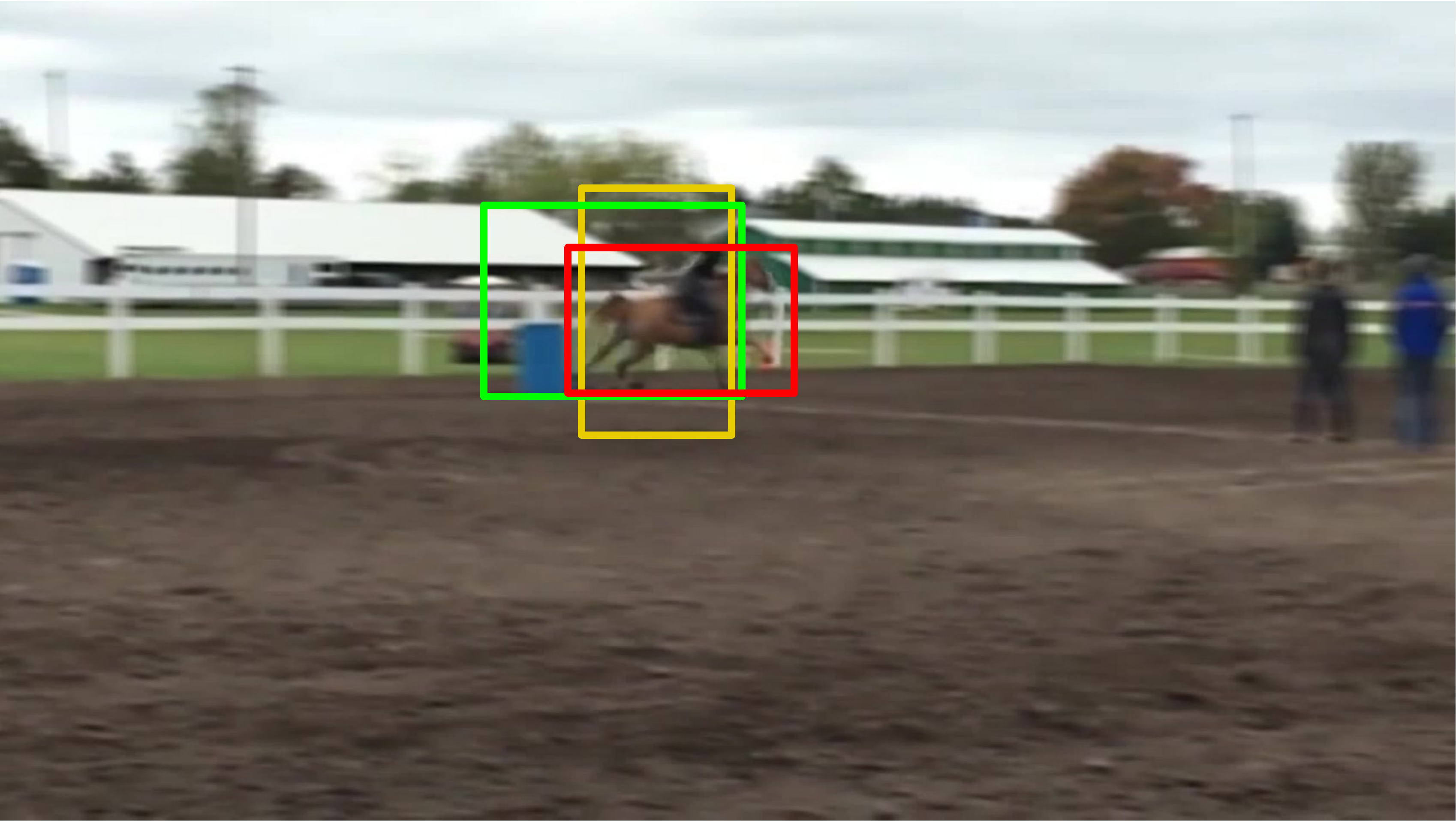}\vspace{-0.5mm}
	\includegraphics*[trim = 2 2 2 2, width = 0.65\columnwidth]{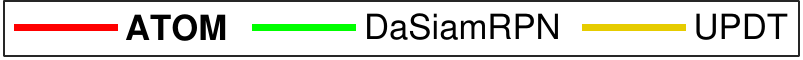}\vspace{-1mm}%
	\caption{A comparison of our approach with state-of-the-art trackers. UPDT \cite{BhatECCV2018}, based on correlation filters, lacks an explicit target state estimation component, performing a brute-force multi-scale search instead. Consequently, it does not handle aspect-ratio changes, which can lead to tracking failure (second row). DaSiamRPN \cite{DaSiamRPN} employs a bounding box regression strategy to estimate the target state, but still struggles in cases of out-of-plane rotation, deformation, etc. Our approach ATOM, employing an overlap prediction network, successfully handles these challenges and provides accurate bounding box predictions.
	}\vspace{-2.5mm}%
	\label{fig:intro}%
\end{figure}

In recent years, the focus of tracking research has been on target classification. Much attention has been invested into constructing robust classifiers, based on e.g.\ correlation filters \cite{DanelljanICCV2015, CSRDCF, DRT}, and exploiting powerful deep feature representations \cite{BhatECCV2018, Valmadre2017cvpr} for this task. On the other hand, target estimation has seen below expected progress. This trend is clearly observed in the recent VOT2018 challenge \cite{VOT2018}, where older trackers such as KCF \cite{Henriques14} and MEEM \cite{MEEM2014} still obtain competitive accuracy while exhibiting vastly inferior robustness. In fact, most current state-of-the-art trackers \cite{BhatECCV2018, DanelljanCVPR2017, DRT} still rely on the classification component for target estimation by performing a multi-scale search. However, this strategy is fundamentally limited since bounding box estimation is inherently a challenging task, requiring high-level understanding of the object's pose (see figure~\ref{fig:intro}). 

In this work, we set out to bridge the performance gap between target classification and estimation in visual object tracking. We introduce a novel tracking architecture consisting of two components designed exclusively for target estimation and classification. Inspired by the recently proposed IoU-Net \cite{IOUNet}, we train the target estimation component to predict the Intersection over Union (IoU) overlap, i.e.\ the Jaccard Index \cite{JaccardIndex}, between the target and an estimated bounding box. Since the original IoU-Net is class-specific, and hence not suitable for generic tracking, we propose a novel architecture for integrating target-specific information into the IoU prediction. We achieve this by introducing a modulation-based network component that incorporates the target appearance in the reference image to obtain target-specific IoU estimates. This further enables our target estimation component to be trained \emph{offline} on large-scale datasets. During tracking, the target bounding box is found by simply maximizing the predicted IoU overlap in each frame.

To develop a seamless and transparent tracking method, we also revisit the problem of target classification with the aim of avoiding unnecessary complexity. 
Our target classification component is simple yet powerful, consisting of a two-layer fully convolutional network head. Unlike our target estimation module, the classification component is trained \emph{online}, providing high robustness against distractor objects in the scene. To ensure real-time performance, we address the problem of efficient online optimization, where gradient descent falls short. Instead, we employ a Conjugate-Gradient-based strategy and demonstrate how it can be easily implemented in modern deep learning frameworks. Our final tracking loop is simple, alternating between target classification, estimation, and model update.

% Results
We perform comprehensive experiments on five challenging benchmarks: NFS \cite{NfS}, UAV123 \cite{UAV123}, TrackingNet \cite{TrackingNet}, LaSOT \cite{LaSOT}, and VOT2018 \cite{VOT2018}. Our tracking approach sets a new state-of-the-art on all five datasets, achieving an absolute gain of $10\%$ on the challenging LaSOT dataset. Moreover, we provide an analysis of our tracker, along with different network architectures for overlap prediction.

\section{Related Work}

In the context of visual tracking, it often makes sense to distinguish between \emph{target classification} and \emph{target estimation} as two separate, but related subtasks. Target classification basically aims at determining the presence of the target object at a certain image location. However, only partial information about the target state is obtained, e.g.\ its image coordinates. Target estimation then aims at finding the full state.
In visual tracking, the target state is often represented by a bounding box, either axis aligned \cite{NfS, OTB2015} or rotated \cite{VOT2018}. State estimation is then reduced to finding the image bounding box that best describes the target in the current frame. 
In the simplest case, the target is rigid and only moves parallel to the camera plane. In such a scenario, \emph{target estimation} reduces to finding the 2D image-location of the target, and therefore does not need to be considered separately from \emph{target classification}. In general, however, objects may undergo radical variations in pose and viewpoint, greatly complicating the task of bounding box estimation.

In the last few years, the challenge of target classification has been successfully addressed by discriminatively training powerful classifiers online \cite{DanelljanICCV2015,Henriques14,MDNet}. In particular, the correlation-based trackers \cite{DanelljanECCV2016,Henriques14,HCF_ICCV15} have gained wide popularity. 
These methods rely on the diagonalizing transformation of circular convolutions, given by the Discrete Fourier Transform, to perform efficient fully convolutional training and inference.
Correlation filter methods often excel at target classification by computing reliable confidence scores in a dense 2D-grid. On the other hand, accurate target estimation has long eluded such approaches. Even finding a one-parameter scale factor has turned out a formidable challenge \cite{danelljan2016discriminative,SAMF} and most approaches resort to the brute-force multi-scale detection strategy with its obvious computational impact. As such, the default method is to apply the classifier alone to perform full state estimation. However, target classifiers are not sensitive to all aspects of the target state, e.g.\ the width and height of the target. In fact, invariance to some aspects of the target state is often considered a valuable property of the discriminative model to improve robustness \cite{SiameseFC,BhatECCV2018,MDNet}. Instead of relying on the classifier, we learn a dedicated target estimation component.

Accurate estimation of an object's bounding box is a complex task, requiring high-level a-priori knowledge. The bounding box depends on the pose and viewpoint of the object, which cannot be modeled as a simple image transformation (e.g.\ uniform image scaling). It is therefore highly challenging, if not impossible, to learn accurate target estimation online from scratch. Many recent methods in the literature have therefore integrated prior knowledge in the form of heavy offline learning \cite{SiamRPN,MDNet,DaSiamRPN}. In particular, SiamRPN \cite{SiamRPN} and its extension \cite{DaSiamRPN} have been shown capable of bounding box regression thanks to extensive offline training. Yet, these Siamese tracking approaches often struggle at the problem of target classification. Unlike, for instance, correlation-based methods, most Siamese trackers do not explicitly account for distractors, since no online learning is performed. While this problem has been partly addressed using simple template update techniques \cite{DaSiamRPN}, it has yet to reach the level of strong online-learned models. In contrast to Siamese methods, we learn the classification model online, while also utilizing extensive offline training for the target estimation task.

\section{Proposed Method}

\begin{figure}[t]
	\centering%
	\newcommand{\wid}{1.0\columnwidth}%
	\includegraphics*[trim = 15 0 8 35, width = \wid]{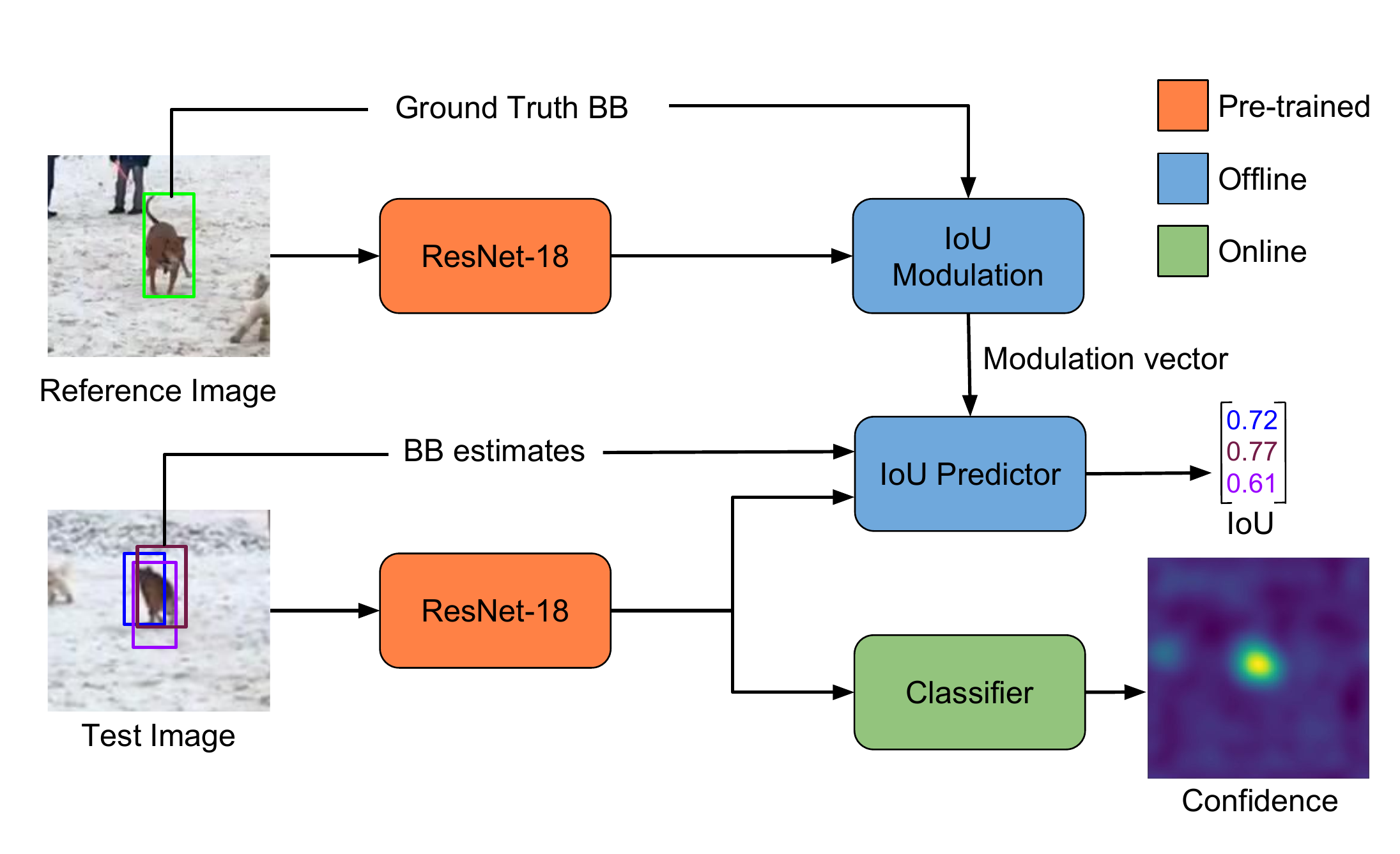}\vspace{-3mm}%
	\caption{Overview of our network architecture for visual tracking. We augment two modules to the pretrained ResNet-18 backbone network (orange). The target estimation module (blue) is trained offline on large-scale datasets to predict the IoU overlap with the target. Using the reference frame and the initial target box, modulation vectors carrying target-specific appearance information are computed. The IoU predictor component then receives features and proposal bounding boxes in the test frame, along with the aforementioned modulation vectors. It estimates the IoU for each input box. The target classification module (green) is trained online to output target confidences in a fully convolutional manner.}%
	\label{fig:method_overview}\vspace{-3mm}%
\end{figure}

In this work, we propose a novel tracking approach consisting of two components: 1) A target estimation module that is learned \emph{offline}; and 2) A target classification module that is learned \emph{online}. That is, following the modern trend in object detection \cite{YOLO9000,FasterRCNN}, we separate the subproblems of target classification and estimation. Yet, both of these tasks are integrated in a unified multi-task network architecture, shown in figure~\ref{fig:method_overview}.

We employ the same backbone network for both the target classification and estimation tasks. For simplicity, we use a ResNet-18 model that is trained on ImageNet and refrain from fine-tuning the backbone in this work. Target estimation is performed by the IoU-predictor network. This network is trained offline on large-scale video tracking and object detection datasets, and its weights are frozen during online tracking. The IoU-predictor takes four inputs: i) backbone features from current frame, ii) bounding box estimates in the current frame, iii) backbone features from a reference frame, iv) the target bounding box in the reference frame. It then outputs the predicted Intersection over Union (IoU) score for each of the current-frame bounding box estimates. During tracking, the final bounding box is obtained by maximizing the IoU score using gradient ascent. The target estimation component is detailed in section~\ref{sec:target-estimation}.

Target classification is performed by another network head. Unlike the target estimation component, the classification network is entirely learned during online tracking. It is exclusively trained to discriminate the target from other objects in the scene by predicting a target confidence score based on backbone features extracted from the current frame. Both training and prediction are performed in a fully convolutional manner to ensure efficiency and coverage. However, training such a network online with conventional approaches, such as stochastic gradient descent, is suboptimal for our online purpose. We therefore propose to use an optimization strategy, based on Conjugate Gradient and Gauss-Newton, that enables fast online training. Moreover, we demonstrate how this approach can be easily implemented in common deep learning frameworks, such as PyTorch, by exploiting the back-propagation functionality. Our target classification approach is described in section~\ref{sec:target-classification} and our final tracking framework is detailed in section~\ref{sec:online-tracking}.

\subsection{Target Estimation by Overlap Maximization}
\label{sec:target-estimation}

\begin{figure*}[t]
	\centering%
	\newcommand{\wid}{0.9\textwidth}%
	\includegraphics*[trim = 0 0 0 0, width = \wid]{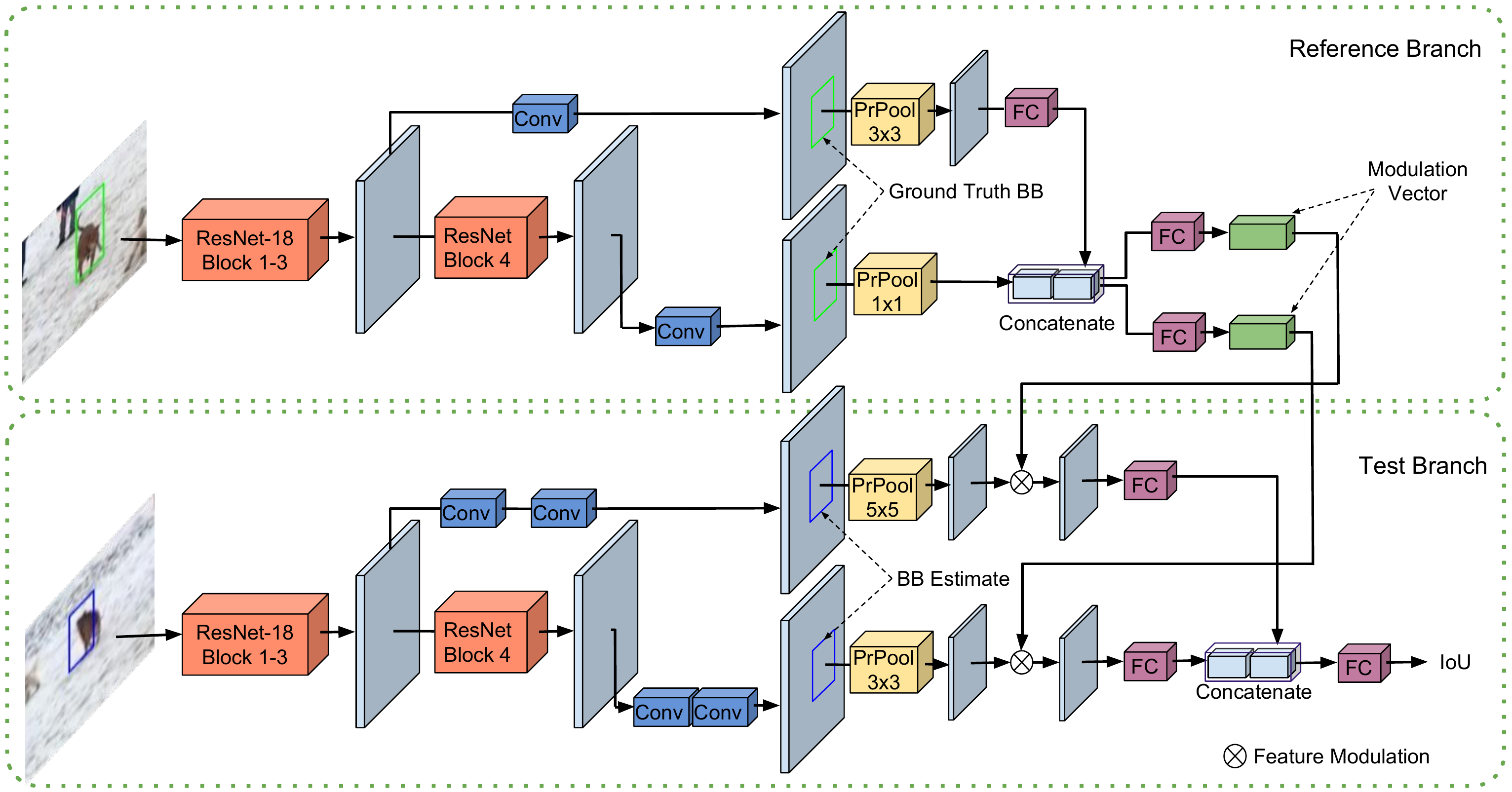}%
	\caption{Full architecture of our target estimation network. ResNet-18 \texttt{Block3} and \texttt{Block4} features extracted from the test image are first passed through two \texttt{Conv} layers. Regions defined by the input bounding boxes are then pooled to a fixed size using \texttt{PrPool} layers. The pooled features are modulated by channel-wise multiplication with the coefficient vector returned by the reference branch. The features are then passed through fully-connected layers to predict the IoU. All \texttt{Conv} and \texttt{FC} layers are followed by \texttt{BatchNorm} and \texttt{ReLU}.
		}%
	\label{fig:iou-network}%
	\vspace{-3mm}
\end{figure*}

In this section, we detail how the target state estimation is performed. The aim of our state estimation component is to determine the target bounding box given a rough initial estimate.
We take inspiration from the IoU-Net \cite{IOUNet}, which was recently proposed for object detection as an alternative to typical anchor-based bounding box regression techniques. In contrast to conventional approaches, the IoU-Net is trained to predict the IoU between an image object and an input bounding box candidate. Bounding box estimation is then performed by maximizing the IoU prediction. 

To describe our target estimation component, we first briefly revisit the IoU-Net model. Given a deep feature representation of an image, $x \in \reals^{W \times H \times D}$, and a bounding box estimate $B \in \reals^4$ of an image object, IoU-Net predicts the IoU between $B$ and the  object. Here $B$ is parametrized as $B=(c_x/w, c_y/h, \log w, \log h)$, where $(c_x, c_y)$ are the image coordinates of the bounding box center. The network uses a Precise ROI Pooling (\verb|PrPool|) \cite{IOUNet} layer to pool the region in $x$ given by $B$, resulting in a feature map $x_\text{B}$ of a pre-determined size. Essentially, \verb|PrPool| is a continuous variant of adaptive average pooling, with the key advantage of being differentiable w.r.t.\ the bounding box coordinates $B$. This allows the bounding box $B$ to be refined by maximizing the IoU w.r.t.\ $B$ through gradient ascent.

\parsection{Network Architecture}
For the task of object detection, independent IoU-Nets are trained in \cite{IOUNet} for each object class.
However, in tracking the target class is generally unknown. Further, unlike object detection, the target is not required to belong to any set of pre-defined classes or be represented in any existing training datasets. Class-specific IoU predictors are thus of little use for generic visual tracking. Instead, \emph{target-specific} IoU predictions are required, by exploiting the target annotation in the first frame. Due to the high-level nature of the IoU prediction task, it is not feasible to train, or even fine-tune the IoU-Net online on a single frame. Thus, we argue that the target estimation network needs to be trained offline to learn a general representation for IoU prediction. 

In the context of visual tracking, where the target object is unknown beforehand, the challenge is thus to construct an IoU prediction architecture that makes effective use of the reference target appearance given at test-time. Our initial experiments showed that naive approaches for fusing the reference image features with the current-frame features yield poor performance (see section~\ref{network_study}). We also found Siamese architectures to provide suboptimal results. In this work, we therefore propose a \emph{modulation-based} network architecture that predicts the IoU for an arbitrary object given only a single reference image. The proposed network is visualized in figure~\ref{fig:iou-network}. Our network has two branches, both of which take backbone features from ResNet-18 \verb|Block3| and \verb|Block4| as input. The \emph{reference branch} inputs features $x_0$ and the target bounding box annotation $B_0$ in the reference image. It returns a modulation vector $c(x_0, B_0)$, consisting of positive coefficients of size $1 \times 1 \times D_z$. As illustrated in figure~\ref{fig:iou-network}, this branch consists of a convolutional layer followed by \verb|PrPool| and a fully connected layer. 

The current image, in which we want to estimate the target bounding box, is processed through the \emph{test branch}. It first extracts a deep representation by feeding the backbone features $x$ through two convolutional layers followed by a \verb|PrPool| with the bounding box estimate $B$. As the test branch extracts general features for IoU prediction, which constitutes a more complex task, it employs more layers and higher pooling resolution compared to the reference branch (see figure \ref{fig:iou-network}). The resulting representation $z(x, B)$ is of size $K \times K \times D_z$, where $K$ is spatial output size of the \verb|PrPool| layer. The computed feature representation of the test image is then \emph{modulated} by the coefficient vector $c$ via a channel-wise multiplication. This creates a target-specific representation for IoU prediction, effectively incorporating the reference appearance information. The modulated representation is finally fed to the IoU predictor module $g$, consisting of three fully connected layers. The predicted IoU of the bounding box $B$ is hence given by
\begin{equation}
	\label{eq:iou-prediction}
	\text{IoU}(B) = g\big(c(x_0, B_0) \cdot z(x, B)\big) \,.
\end{equation}
To train the network, we minimize the prediction error of \eqref{eq:iou-prediction}, given annotated data. During tracking we maximize \eqref{eq:iou-prediction} w.r.t.\ $B$ to estimate the target state.

\parsection{Training}
From \eqref{eq:iou-prediction} it is clear that the entire IoU prediction network can be trained in an end-to-end fashion, using bounding-box-annotated image pairs. 
We use the training splits of the recently introduced Large-scale Single Object Tracking (LaSOT) dataset \cite{LaSOT} and TrackingNet \cite{TrackingNet}. We sample image pairs from the videos with a maximum gap of 50 frames. Similar to \cite{DaSiamRPN}, we augment our training data with synthetic image pairs from the COCO dataset \cite{COCO} to have more diverse classes. From the reference image, we sample a square patch centered at the target, with an area of about $5^2$ times the target area. From the test image, we sample a similar patch, with some perturbation in the position and scale to simulate the tracking scenario. These cropped regions are then resized to a fixed size. For each image pair we generate 16 candidate bounding boxes by adding Gaussian noise to the ground truth coordinates, while ensuring a minimum IoU of $0.1$. We use image flipping and color jittering for data augmentation. As in \cite{IOUNet}, the IoU is normalized to $[-1, 1]$.

The weights in our head network are initialized using \cite{KaimingInit}. For the backbone network, we freeze all weights during training.  We use the mean-squared error loss function and train for $40$ epochs with $64$ image pairs per batch. The ADAM \cite{ADAM} optimizer is employed with initial learning rate of $10^{-3}$, and using a factor $0.2$ decay every $15$ epochs.

\subsection{Target Classification by Fast Online Learning}
\label{sec:target-classification}

While the target estimation module provides accurate bounding box outputs, it lacks the ability to robustly discriminate between the target object and background distractors. We therefore complement the estimation module with a second network head, whose sole purpose is to perform this discrimination. Unlike the estimation component, the target classification module is exclusively trained online, to predict a target confidence score. Since the goal of the target classification module is to provide a rough 2D-location of the object, we wish it to be \emph{invariant} to the size and scale of the target. Instead, it should emphasize robustness by minimizing false detections.

\parsection{Model}
Our target classification module is a 2-layer fully convolutional neural network, formally defined as
\begin{equation}
	\label{eq:classification-net}
	f(x; w) = \phi_2(w_2 \conv \phi_1(w_1 \conv x)) \,.
\end{equation}
Here, $x$ is the backbone feature map, $w = \{w_1, w_2\}$ are the parameters of the network, $\phi_1, \phi_2$ are activation functions and $\conv$ denotes standard multi-channel convolution. While our framework is general, allowing more complex models for this purpose, we found such a simple model sufficient and beneficial in terms of computational efficiency.

Inspired by the recent success of discriminative correlation filter (DCF) approaches, we formulate a similar learning objective based on the $L^2$ classification error,
\begin{equation}
	\label{eq:classification-loss}
	L(w) = \sum_{j=1}^{m} \gamma_j \|f(x_j; w) - y_j\|^2 + \sum_k \lambda_k \|w_k\|^2 \,.
\end{equation}
Each training sample feature map $x_j$ is annotated by the classification confidences $y_j \in \reals^{W \times H}$, set to a sampled Gaussian function centered at the target location. The impact of each training sample is controlled by the weight $\gamma_j$, while the amount of regularization on $w_k$ is set by $\lambda_k$.

\parsection{Online Learning}
\newcommand{\dw}{\Delta w}
A brute-force approach to minimize \eqref{eq:classification-loss} would be to apply standard gradient descent or its stochastic twin. These approaches are easily implemented in modern deep learning libraries, but are not well suited for online learning due to their slow convergence rates. We therefore develop a more sophisticated optimization strategy that is tailored for such online learning problems, yet requiring only little added implementation complexity. First, we define the residuals of the problem as $r_j(w) = \sqrt{\gamma_j}(f(x_j; w) - y_j)$ for $j \in \{1,\ldots,m\}$ and $r_{m+k}(w) = \sqrt{\lambda_k} w_k$ for $k = 1, 2$. The loss \eqref{eq:classification-loss} is then equivalently written as the squared $L^2$ norm of the residual vector $L(w) = \|r(w)\|^2$, where $r(w)$ is the concatenation of all residuals $r_j(w)$. We utilize the quadratic Gauss-Newton approximation $\tilde{L}_w(\dw) \approx L(w+\dw)$, obtained from a first order Taylor expansion of the residuals $r(w+\dw) \approx r_w + J_w \dw$ at the current estimate $w$,
\begin{equation}
	\label{eq:GN-loss}
	\tilde{L}_w(\dw) = \dw\tp J_w\tp J_w \dw + 2 \dw\tp J_w\tp r_w + r_w\tp r_w \,.
\end{equation}
Here, we have defined $r_w = r(w)$ and $J_w = \frac{\partial r}{\partial w}$ is the Jacobian of $r$ at $w$. The new variable $\dw$ represents the increment in the parameters $w$. 

The Gauss-Newton subproblem \eqref{eq:GN-loss} forms a positive definite quadratic function, allowing the deployment of specialized machinery such as the Conjugate Gradient (CG) method. While a full description of CG is outside the scope of this paper (see \cite{CGpain} for a full treatment), intuitively it finds an optimal search direction $p$ and step length $\alpha$ in each iteration.
Since the CG algorithm consists of simple vector operations, it can be implemented with only a few lines of python code. The only challenging aspect of CG is the evaluation of the operation $J_w\tp J_w p$ for a search direction $p$.

We note that CG has been successfully deployed in some DCF tracking approaches \cite{DanelljanCVPR2017,DanelljanECCV2016,DRT}. However, these implementations rely on hand-coding all operations in order to implement $J_w\tp J_w p$, requiring much tedious work and derivations even for a simple model \eqref{eq:classification-net}. This approach also lacks flexibility since any minor modification of the architecture \eqref{eq:classification-net}, such as adding a layer or changing a non-linearity, may require comprehensive re-derivation and implementation work. In this paper, we therefore demonstrate how to implement CG for \eqref{eq:GN-loss} by exploiting the backpropagation functionality of modern deep learning frameworks, such as PyTorch. Our implementation only requires the user to supply the function $r(w)$ for evaluating the residuals, which is easy to implement. Our algorithm is therefore applicable to any shallow learning problem of the form \eqref{eq:classification-loss}.

\newcommand{\bp}{\mathtt{BackProp}}

To find a strategy for evaluating $J_w\tp J_w p$, we first consider a vector $u$ of the same size as the residuals $r(w)$. By computing the gradient of their inner product, we obtain $\frac{\partial}{\partial w} (r(w)\tp u) = \frac{\partial r}{\partial w}\tp u = J_w\tp u$. In fact, this is the standard operation of the backpropagation procedure, namely to apply the transposed Jacobian at each node in the computational graph, starting at the output. We can thus define backpropagation of a \emph{scalar} function $s$ with respect to a variable $v$ as $\bp(s, v) = \frac{\partial s}{\partial v}$. Now, as shown above, we have $\bp(r\tp u, w) = J_w\tp u$. However, this only accounts for the second product in $J_w\tp J_w p$. We first have to compute $J_w p$, which involves the application of the Jacobian itself (not its transpose). Thankfully, the Jacobian of the function $u \mapsto J_w\tp u$ is trivially $J_w\tp$, since the function is linear. We can therefore transpose it by applying backpropagation. By letting $h \defeq J_w\tp u = \bp(r\tp u, w)$, we get $J_w p = \frac{\partial}{\partial u}(h\tp p) = \bp(h\tp p, u)$.

Given the above mentioned result, we outline the entire optimization procedure in algorithm \ref{alg:GNCG}. It applies $N_\text{GN}$ Gauss-Newton iterations, each encompassing $N_\text{CG}$ Conjugate Gradient iterations for minimizing the resulting subproblem \eqref{eq:GN-loss}. Each CG iteration requires two $\bp$ calls for evaluating $q_1 = J_w p$ and $q_2 = J_w\tp q_1$, respectively. There is a need for computing $h = J_w\tp u$ once in the outer loop. Note that in each call to $\bp$ in algorithm \ref{alg:GNCG}, one of the vectors in the inner product is treated as constant, i.e.\ gradients are not propagated through it. This is highlighted as comments in algorithm~\ref{alg:GNCG} for clarity. It is noteworthy that the optimization algorithm is virtually parameter free, only the number of iterations need to be set. In comparison to gradient descent, the CG-based method adaptively computes the learning rate $\alpha$ and momentum $\beta$ in each iteration. Observe that $g$ is the negative gradient of \eqref{eq:GN-loss}. 

\newcommand{\assign}{\leftarrow}
\newcommand{\algcomment}[1]{\hspace{3mm}{\footnotesize\# #1}}
\newcommand{\sep}{\,,\quad}
\begin{algorithm}[t]
	\caption{Classification component optimization.}
	\begin{algorithmic}[1]
		\Require Net weights $w$, residual function $r(w)$, $N_\text{GN}$, $N_\text{CG}$
		\For{$i = 1, \ldots, N_\text{GN}$}
			\State $r \assign r(w) \sep u \assign r$
			\State $h \assign \bp(r\tp u, w)$ \hspace{7.7mm}\algcomment{Treat $u$ as constant}
			\State $g \assign -h \sep p \assign 0 \sep \rho_1 \assign 1 \sep \dw \assign 0$
			\For{$n = 1, \ldots, N_\text{CG}$}
				\State $\rho_2 \assign \rho_1 \sep \rho_1 \assign g\tp g \sep \beta \assign \frac{\rho_1}{\rho_2}$
				\State $p \assign g + \beta p$
				\State $q_1 \assign \bp(h\tp p, u)$ \hspace{1.7mm}\algcomment{Treat $p$ as constant}
				\State $q_2 \assign \bp(r\tp q_1, w)$ \hspace{0mm}\algcomment{Treat $q_1$ as constant}
				\State $\alpha \assign \frac{\rho_1}{q_2\tp p}$
				\State $g \assign g - \alpha q_2$
				\State $\dw \assign \dw + \alpha p$
			\EndFor
			\State $w \assign w + \dw$
		\EndFor
	\end{algorithmic}
	\label{alg:GNCG}
\end{algorithm}

\subsection{Online Tracking Approach}
\label{sec:online-tracking}
Our tracker \textbf{ATOM} is implemented in Python, using PyTorch. It runs at over $30$ FPS on an Nvidia GT-1080 GPU.

\parsection{Feature extraction}
We use ResNet-18 pretrained on ImageNet as our backbone network. For target classification, we employ block 4 features, while the target estimation component uses both block 3 and 4 as input. Features are always extracted from patches of size $288 \times 288$ from image regions corresponding to 5 times the estimated target size. Note that ResNet-18 feature extraction is shared and only performed on a single image patch every frame.

\parsection{Classification Model} 
The first layer in our classification head \eqref{eq:classification-net} consists of a $1\times 1$ convolutional layer $w_1$, which reduces the feature dimensionality to 64. As in \cite{DanelljanCVPR2017}, the purpose of this layer is to limit memory and computational requirements. The second layer employs a $4 \times 4$ kernel $w_2$ with a single output channel. We set $\phi_1$ to identity since we did not observe any benefit of using a non-linearity at this layer. We use a continuously differentiable parametric exponential linear unit (PELU) \cite{PELU} as output activation: $\phi_2(t) = t, t \geq 0$ and $\phi_2(t) = \alpha (e^{\frac{t}{\alpha}} - 1), t \leq 0$. Setting $\alpha = 0.05$ allows us to ignore easy negative examples in the loss \eqref{eq:classification-loss}. We found the continuous differentiability of $\phi_2$ to be advantageous for optimization.

In the first frame, we perform data augmentation by applying varying degrees of translation, rotation, blur, and dropout, similar to \cite{BhatECCV2018}, resulting in 30 initial training samples $x_j$. We then apply algorithm~\ref{alg:GNCG} with $N_\text{GN} = 6$ and $N_\text{CG} = 10$ to optimize the parameters $w$. Subsequently, we only optimize the final layer $w_2$, using $N_\text{GN} = 1$ and $N_\text{CG} = 5$ every 10th frame. In every frame, we add the extracted feature map $x_j$ as a training sample, annotated by a Gaussian $y_j$ centered at the estimated target location. The weights $\gamma_j$ in \eqref{eq:classification-loss} are updated with a learning rate of $0.01$.

\parsection{Target Estimation}
We first extract features at the previously estimated target location and scale. We then apply the classification model \eqref{eq:classification-net} and find the 2D-position with the maximum confidence score. Together with the previously estimated target width and height, this generates the initial bounding box $B$. While it is possible to perform state estimation using this single proposal, we found that local maxima are better avoided using multiple random initializations. We therefore generate a set of $10$ initial proposals by adding uniform random noise to $B$. The predicted IoU \eqref{eq:iou-prediction} of each box is maximized using $5$ gradient ascent iterations with a step length of $1$. The final prediction is obtained by taking the mean of the $3$ bounding boxes with highest IoU. No further post-processing or filtering, as in e.g.\ \cite{SiamRPN} is performed. This refined state also annotates the training sample $(x_j,y_j)$, as described earlier. Note that the modulation vector $c(x_0, B_0)$ in \eqref{eq:iou-prediction} is precomputed in the first frame.

\parsection{Hard Negative Mining}
To further robustify our classification component in the presence of distractors, we adopt a hard negative mining strategy, common in many visual trackers \cite{MDNet,DaSiamRPN}. If a distractor peak is detected in the classification scores, we double the learning rate of this training sample and instantly run a round of optimization with standard settings ($N_\text{GN} = 1, N_\text{CG} = 5$). We also determine the target as lost if the score falls below $0.25$. While the hard negative strategy is not fundamental to our framework, it provides some additional robustness (see section~\ref{method_ablation}). 

\section{Experiments}

We evaluate the proposed tracker \textbf{ATOM} on five benchmarks: Need for Speed (NFS) \cite{NfS}, UAV123 \cite{UAV123}, TrackingNet \cite{TrackingNet}, LaSOT \cite{LaSOT}, and VOT2018 \cite{VOT2018}. Detailed results are provided in the supplementary material.

\subsection{IoU Prediction Architecture Analysis}
\label{network_study}
Here, we study the impact of various architectural choices for the IoU prediction module, presented in section~\ref{sec:target-estimation}. Our analysis is performed on the combined UAV123 \cite{UAV123} and NFS ($30$ FPS version)~\cite{NfS} datasets, summing to $223$ videos. These datasets contain a high variety of videos that are challenging in many aspects, such as deformation, view change, occlusion, fast motion and distractors. We evaluate the trackers based on the overlap precision metric ($\text{OP}_{T}$), defined as the percentage of frames having bounding box IoU overlap larger than a threshold $T$ with the ground truth. We also report the area-under-the-curve (AUC) score, defined as $\text{AUC}=\int_{0}^{1}\text{OP}_{T}\diff{T}$. In all experiments, we report the average result over $5$ runs.

\begin{table}[!b]
	\centering\vspace{-1mm}
	\resizebox{1.01\columnwidth}{!}{%
		\begin{tabular}{l@{~}c@{~~}c@{~~}c@{~~}c@{~~}c@{~~}c@{~~}}
\toprule
&Baseline&Modulation&Concatenation&Siamese&Modulation&Modulation\\
&(Block 3\&4)&(Block 3\&4)  &(Block 3\&4)&(Block 3\&4)&(Block 3)&(Block 4)\\\midrule
$\text{OP}_{0.50} (\%)$&68.3&\textbf{76.3}&67.5&75.1&73.4&73.6\\
$\text{OP}_{0.75} (\%)$&38.6&\textbf{48.4}&37.9&47.6&44.5&38.9\\
AUC (\%)&56.7&\textbf{62.3}&56.3&61.7&60.3&58.5\\\bottomrule
\end{tabular}

	}\vspace{1mm}%
	\caption{Analysis of different architectures for IoU prediction on the combined NFS and UAV123 datasets. For each method, we indicate in parenthesis the backbone feature layers that are used as input. The baseline approach, which does not employ a reference branch to integrate target specific information, provides poor results. Among the different architectures, the modulation based approach, using both block 3 and 4, achieves the best results.}
	\label{tab:iou_ablation}%
	\vspace{-1mm}
\end{table}

\parsection{Reference image} We compare with a baseline approach that excludes target specific information by removing the reference branch in our architecture. That is, the baseline network only uses the test frame to predict the IoU. The results of this investigation are shown in table \ref{tab:iou_ablation}. Excluding the reference frame deteriorates the results by over $5.5\%$ AUC score. This demonstrates the importance of exploiting target-specific appearance information in order to accurately predict the IoU for an arbitrary object.

\parsection{Integration of target appearance} We investigate different network architectures for integrating the reference image features for IoU prediction. We compare our feature modulation based method, presented in section~\ref{sec:target-estimation}, with two alternative architectures. \textbf{Concatenation:} Activations from the reference and test branches are concatenated before the final IoU prediction layers. \textbf{Siamese:} Using identical architecture for both branches and performing final IoU prediction as a scalar product of their outputs. All the networks are trained using the same setup, with ResNet18 \verb|Block3| and \verb|Block4| features as input. For a fair comparison, we ensure that all networks have the same depth and similar number of trainable parameters. Results are shown in table~\ref{tab:iou_ablation}. Naively concatenating the features from the reference image and the test image achieves an AUC of $56.3\%$. Our Siamese-based architecture obtains better results, with an AUC of $61.7\%$ and $\text{OP}_{0.50}$ of $75.1\%$. Our modulation-based method further improves the results, giving an absolute gain of $1.2\%$ in $\text{OP}_{0.50}$ and achieves an AUC of $62.3\%$.

\parsection{Backbone feature layers} We evaluate the impact of using different feature blocks from the backbone ResNet-18 (table \ref{tab:iou_ablation}). Using features from only \verb|Block3| leads to an AUC of $60.3\%$, while only \verb|Block4| gives an AUC of $58.5\%$. Fusing features from both the blocks leads to a significant improvement, giving an AUC score of $62.3\%$. This indicates that \verb|Block3| and \verb|Block4| features have complementary information useful for predicting the IoU.

\subsection{Ablation Study}
\label{method_ablation}
We perform an ablation study to demonstrate the impact of each component in the proposed method. We use the same dataset and the evaluation criteria as in section~\ref{network_study}.

\begin{table}[t]
	\centering
	\resizebox{0.85\columnwidth}{!}{%
		\begin{tabular}{l@{~~~}c@{~~~}c@{~~~}c@{~~~}c@{~~~}c@{~~~}c@{~~~}}
\toprule
&\textbf{ATOM}&Multi-Scale&No Classif.&GD&GD++&No HN\\\midrule
$\text{OP}_{0.50} (\%)$&\textbf{76.3}&66.2&52.3&74.5&74.8&75.9\\
$\text{OP}_{0.75} (\%)$&\textbf{48.4}&26.0&35.1&47.4&47.3&48.1\\
AUC (\%)&\textbf{62.3}&53.7&43.0&60.9&61.1&61.9\\\bottomrule
\end{tabular}

	}\vspace{1mm}%
	\caption{Impact of each component in the proposed approach on the combined NFS and UAV123 datasets. We compare the target estimation component with the brute-force multi-scale approach and analyze the impact of our classification module, online optimization strategy, and hard-negative mining scheme.}
	\label{tab:method_ablation}%
	\vspace{-3mm}
\end{table}

\parsection{Target Estimation}
We compare our target state estimation component, presented in section~\ref{sec:target-estimation}, with a brute-force \textbf{multi-scale} search approach employing \emph{only} the classification model. This approach mimics the common practice in correlation filter based methods, extracting features at $5$ scales with a scale ratio of $1.02$. The classification component is then evaluated on all scales, selecting the location and scale with the highest confidence score as the new target state. Results are shown in table \ref{tab:method_ablation}. Our approach significantly outperforms the multi-scale method by $8.6\%$ in AUC. Further, our approach almost doubles the percentage of highly accurate bounding box predictions, as measured by $\text{OP}_{0.75}$. These results highlight the importance of treating target state estimation as a high-level visual task.

\parsection{Target Classification}
We investigate the impact of the target classification component (section \ref{sec:target-classification}) by excluding it from our tracking framework. 
\textbf{No Classif} in table \ref{tab:method_ablation} only employs the target estimation module for tracking, using a larger search region. The resulting method achieves an AUC of $43.0\%$, almost $20\%$ less than our approach.

\parsection{Online Optimization}
We investigate the impact of the optimization strategy presented in algorithm~\ref{alg:GNCG}, by comparing it with gradient descent. We use carefully tuned learning rate and momentum parameters for the gradient descent approach. In the version termed \textbf{GD}, we run the same number of $\mathtt{BackProp}$ operations as in our algorithm, obtaining the same speed as of our tracker. We also compare with \textbf{GD++}, running $5$ times as many iterations as in \textbf{GD}, thus running at significantly slower frame rates. In both cases, the proposed Gauss-Newton approach outperforms gradient descent by more than $1.2\%$ AUC score (Table \ref{tab:method_ablation}). Note that even a 5-fold increase of iterations does not provide any significant improvement (only $0.2\%$), indicating slow convergence.

\parsection{Hard Negative Mining} We evaluate our method without the \textbf{H}ard \textbf{N}egative mining component (section \ref{sec:online-tracking}), resulting in an AUC of $61.9\%$. This suggests the hard negative mining adds some robustness ($0.4\%$ AUC) to our tracker.

\begin{figure}[t]
	\newcommand{\wid}{0.5\columnwidth}%
	\centering\vspace{-5mm}%
	\subfloat[NFS\label{fig:sota_nfs}]{\includegraphics[width = \wid]{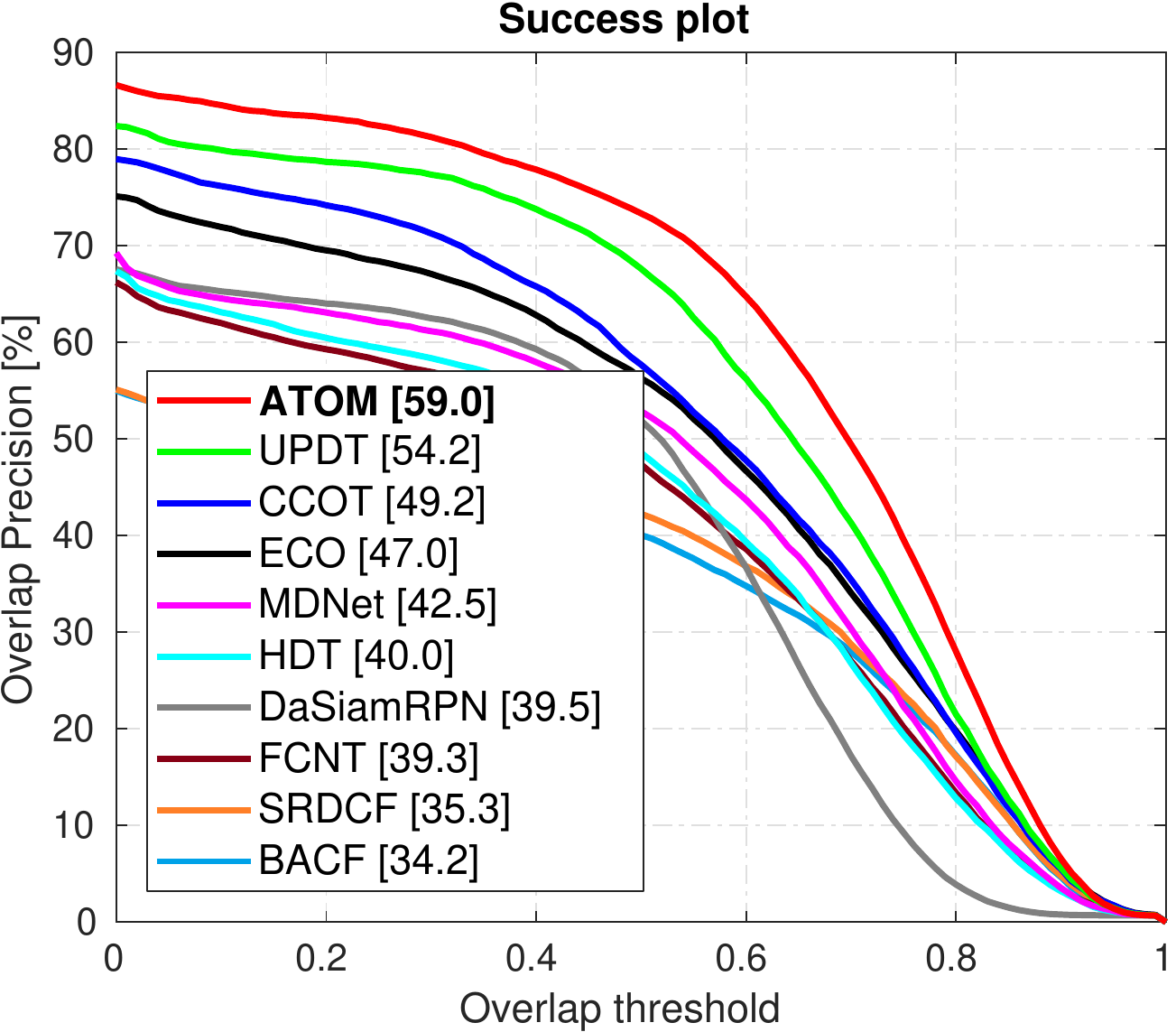}}%
	\subfloat[UAV123\label{fig:sota_uav}]{\includegraphics[width = \wid]{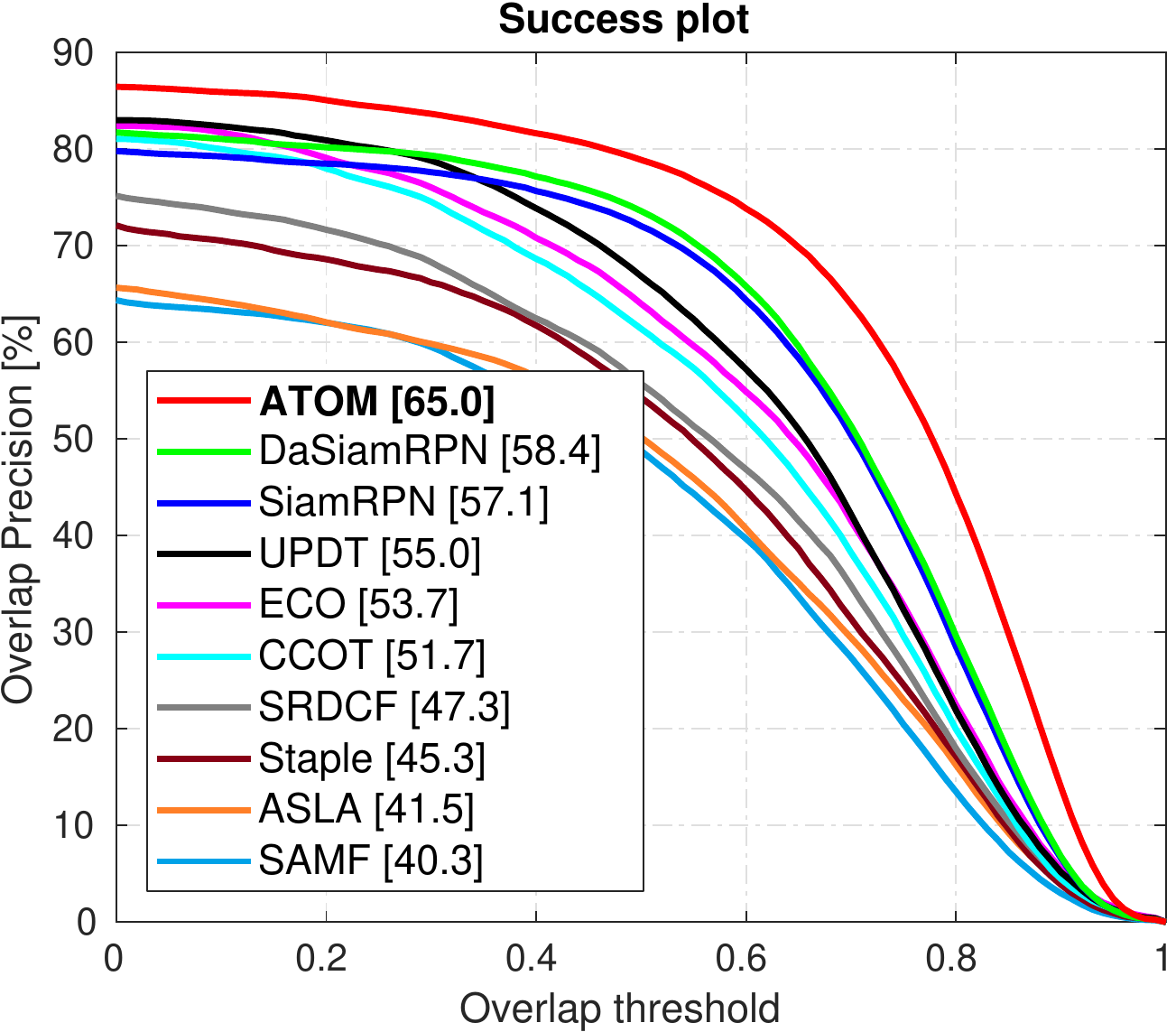}}%
	\caption{Success plots on NFS (a) and UAV123 (b). In both cases, our approach improves the state-of-the-art by a large margin.}%
	\label{fig:sotanfsuav}\vspace{-1mm}
\end{figure}

\subsection{State-of-the-art Comparison}
\label{sec:sota}%
We present the comparison of our tracker with state-of-the-art methods on five challenging tracking datasets.

\begin{table}[t]
	\centering
	\resizebox{\columnwidth}{!}{%
		\begin{tabular}{l@{~}c@{~~}c@{~~}c@{~~}c@{~~}c@{~~}c@{~~}c@{~~}c@{~~}c@{~~}c@{~~}}
\toprule
&Staple&SAMF&CSRDCF&ECO&DaSiam-&SiamFC&CFNet&MDNet&UPDT&\textbf{ATOM}\\
&\cite{Staple}&\cite{SAMF}&\cite{CSRDCF}&\cite{DanelljanCVPR2017}&RPN \cite{DaSiamRPN}&\cite{SiameseFC}&\cite{Valmadre2017cvpr}&\cite{MDNet}&\cite{BhatECCV2018}\\\midrule
Precision (\%)&47.0&47.7&48.0&49.2&41.3&53.3&53.3&\textbf{\textcolor{blue}{56.5}}&55.7&\textbf{\textcolor{red}{64.8}}\\
Norm.\ Prec. (\%)&60.3&59.8&62.2&61.8&60.2&66.6&65.4&\textbf{\textcolor{blue}{70.5}}&70.2&\textbf{\textcolor{red}{77.1}}\\
Success (\%)&52.8&50.4&53.4&55.4&56.8&57.1&57.8&60.6&\textbf{\textcolor{blue}{61.1}}&\textbf{\textcolor{red}{70.3}}\\\bottomrule
\end{tabular}

	}\vspace{1mm}%
	\caption{State-of-the-art comparison on the TrackingNet test set in terms of precision, normalized precision, and success. Our approach significantly outperforms UPDT, achieving a relative gain of $15\%$ in terms of success.}%
	\label{tab:trackingnet_sota}%
	\vspace{-2mm}
\end{table}

\parsection{Need For Speed \cite{NfS}} We evaluate on the $30$ FPS version of the dataset. Figure~\ref{fig:sota_nfs} shows the success plot over all the 100 videos, reporting AUC scores in the legend. CCOT~\cite{DanelljanECCV2016} and UPDT \cite{BhatECCV2018}, both based on correlation filters, achieve AUC scores of $49.2\%$ and $54.2\%$ respectively. Our tracker significantly outperforms UPDT with a relative gain of $9\%$.

\parsection{UAV123 \cite{UAV123}} Figure~\ref{fig:sota_uav} displays the success plot over all the 123 videos. DaSiamRPN~\cite{DaSiamRPN} and its predecessor SiamRPN~\cite{SiamRPN} employ a target estimation component based on bounding box regression. Compared to other approaches, DaSiamRPN achieves a superior AUC of $58.4\%$, owing to its accuracy. Our tracker, employing an overlap maximization strategy for target estimation, significantly outperforms DaSiamRPN by achieving an AUC of $65.0\%$.

\parsection{TrackingNet \cite{TrackingNet}} This is a recently introduced large-scale dataset consisting of real-world videos sampled from YouTube. The trackers are evaluated using an online evaluation server on a test set of $511$ videos. Table~\ref{tab:trackingnet_sota} shows the results in terms of precision, normalized precision, and success. In terms of precision and success, MDNet~\cite{MDNet} achieves scores of $56.5\%$ and $60.6\%$ respectively. Our tracker outperforms MDNet with relative gains of $14\%$ and $16\%$ in terms of precision and success respectively.

\begin{table}[!t]
	\centering\vspace{-1mm}
	\resizebox{1.01\columnwidth}{!}{%
		\begin{tabular}{l@{~}c@{~~}c@{~~}c@{~~}c@{~~}c@{~~}c@{~~}c@{~~}c@{~~}c@{~~}c@{~~}}
\toprule
&STRCF&SINT&ECO&DSiam&StructSiam&SiamFC&VITAL&MDNet&DaSiam-&\textbf{ATOM}\\
&\cite{STRCF}&\cite{Tao2016Sint}&\cite{DanelljanCVPR2017}&\cite{Gua2017DSiam}&\cite{StructSiam}&\cite{SiameseFC}&\cite{VITAL}&\cite{MDNet}&RPN\cite{DaSiamRPN}&\\\midrule
Norm.\ Prec. (\%)&34.0&35.4&33.8&40.5&41.8&42.0&45.3&46.0&\textbf{\textcolor{blue}{49.6}}&\textbf{\textcolor{red}{57.6}}\\
Success (\%)&30.8&31.4&32.4&33.3&33.5&33.6&39.0&39.7&\textbf{\textcolor{blue}{41.5}}&\textbf{\textcolor{red}{51.5}}\\\bottomrule
\end{tabular}

	}\vspace{1mm}%
	\caption{State-of-the-art comparison on the LaSOT dataset in terms of normalized precision and success.}
	\label{tab:lasot}%
	\vspace{-1mm}
\end{table}

\parsection{LaSOT \cite{LaSOT}} We evaluate our approach on the test split consisting of 280 videos. Table \ref{tab:lasot} shows the results in terms of normalized precision and success. Among previous approaches, DaSiamRPN achieves the best success scores. Our approach significantly outperforms DaSiamRPN with an absolute gain of $10.0\%$ in success.

\parsection{VOT2018 \cite{VOT2018}} This dataset consists of 60 videos and the performance is evaluated in terms of robustness (failure rate) and accuracy (average overlap in the course of successful tracking). The two measures are merged in a single metric, Expected Average Overlap (EAO), which provides the overall performance ranking. Table~\ref{tab:vot_sota} shows the comparison of our approach with the top-10 trackers in the VOT2018 competition \cite{VOT2018}. Among the top trackers, only DaSiamRPN uses an explicit target state estimation component, achieving higher accuracy compared to its DCF-based counterparts like LADCF~\cite{LADCF} and MFT. Our approach ATOM achieves the best accuracy, while having competitive robustness. Further, our tracker obtains the best EAO score of $0.401$, with a relative gain of $3\%$ over LADCF.

\begin{table}[t]
	\centering
	\resizebox{\columnwidth}{!}{%
		\begin{tabular}{l@{~}c@{~~}c@{~~}c@{~~}c@{~~}c@{~~}c@{~~}c@{~~}c@{~~}c@{~~}c@{~~}c@{~~}}
\toprule
&DLSTpp&SASiamR&CPT&DeepSTRCF&DRT&RCO&UPDT&DaSiam-&MFT&LADCF&\textbf{ATOM}\\
&\cite{VOT2018}&\cite{SASiamR}&\cite{VOT2018}&\cite{STRCF}&\cite{DRT}&\cite{VOT2018}&\cite{BhatECCV2018}&RPN \cite{DaSiamRPN}&\cite{VOT2018}&\cite{LADCF}\\\midrule
EAO&0.325&0.337&0.339&0.345&0.356&0.376&0.378&0.383&0.385&\textbf{\textcolor{blue}{0.389}}&\textbf{\textcolor{red}{0.401}}\\
Robustness&0.224&0.258&0.239&0.215&0.201&\textbf{\textcolor{blue}{0.155}}&0.184&0.276&\textbf{\textcolor{red}{0.140}}&0.159&0.204\\
Accuracy&0.543&0.566&0.506&0.523&0.519&0.507&0.536&\textbf{\textcolor{blue}{0.586}}&0.505&0.503&\textbf{\textcolor{red}{0.590}}\\\bottomrule
\end{tabular}

	}\vspace{1mm}%
	\caption{State-of-the-art comparison on the public VOT2018 dataset in terms of expected average overlap (EAO), robustness (tracking failure), and accuracy. Our tracker outperforms all the previous methods in terms of EAO.}%
	\label{tab:vot_sota}%
	\vspace{-3mm}
\end{table}

\section{Conclusions} 
We propose a novel tracking architecture with explicit components for target estimation and classification. The estimation component is trained offline on large-scale datasets to predict the IoU overlap between the target and a bounding box estimate. Our architecture integrates target-specific knowledge by performing feature modulation. The classification component consists of a two-layer fully convolutional network head and is trained online using a dedicated optimization approach. Comprehensive experiments are performed on four tracking benchmarks. Our approach provides accurate target estimation while being robust against distractor objects in the scene, outperforming previous methods on all four datasets.

\noindent\textbf{Acknowledgments}:
This work was supported by  SSF (SymbiCloud), Swedish Research Council (EMC${}^2$, grant 2018-04673), ELLIIT, and WASP.

{\small
\bibliographystyle{ieee}
\bibliography{references}
}

\clearpage
\setcounter{equation}{0}
\setcounter{figure}{0}
\setcounter{table}{0}
\setcounter{section}{0}

\renewcommand{\theequation}{S\arabic{equation}}
\renewcommand{\thefigure}{S\arabic{figure}}
\renewcommand{\thetable}{S\arabic{table}}
\renewcommand{\thesection}{S\arabic{section}}

\begin{center}
	\textbf{\large Supplementary Material}
\end{center}

	In this supplementary material we provide additional details and results. Section \ref{sec:network_architecures} provides details about the other network architectures evaluated for IoU prediction in section \ref{network_study} of the main paper. Section~\ref{sec:convergence} performs an empirical convergence analysis of the employed optimization procedure and the gradient descent. Detailed results on the LaSOT \cite{LaSOT} dataset are provided in section~\ref{sec:lasot}. Section~\ref{sec:otb} provides results on the OTB-100~\cite{OTB2015} dataset. The impact of the training data on performance of our tracker is analyzed in section~\ref{sec:training_data}. Section \ref{sec:additional_results} provides detailed results on the UAV123 dataset. A video showing qualitative results of our tracker can be found at \url{https://youtu.be/T8x8i1KkYGk}.

	\section{Network Architectures for IoU Prediction}
	\label{sec:network_architecures}
\begin{figure}[t]
		\centering
		\includegraphics[width=\columnwidth,trim={0.0cm 0.0cm 0.0cm 0.0cm},clip]{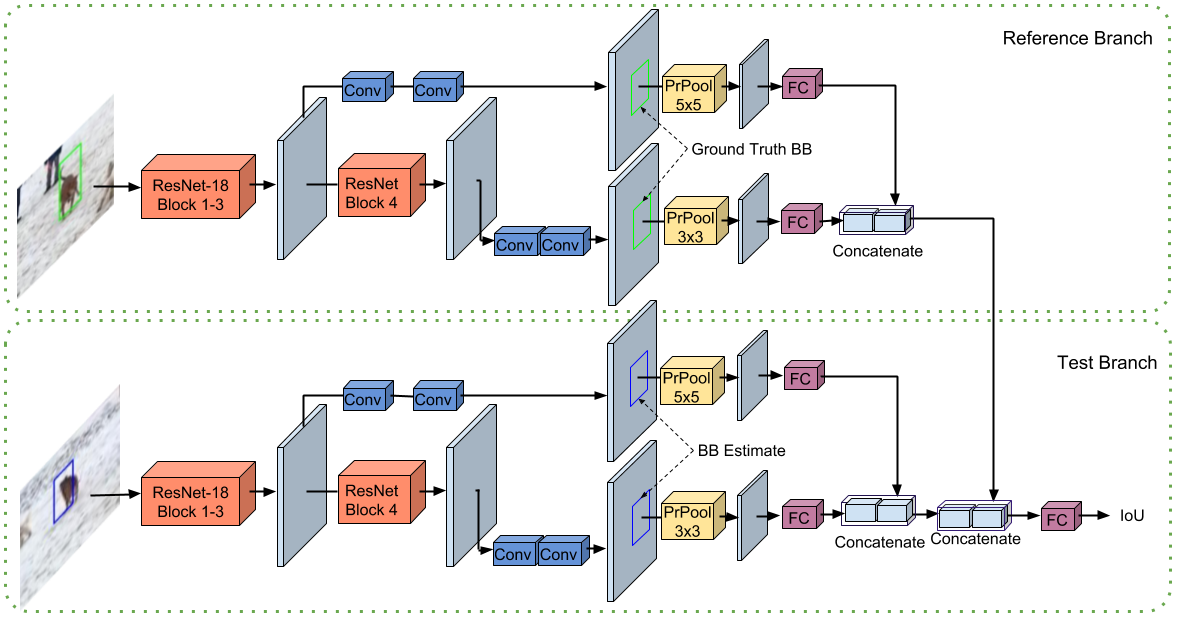}\vspace{-1mm}
		\vspace{-3mm}
		\caption{Architecture of the \textbf{Concatenation} network for IoU prediction evaluated in section \ref{network_study} in the paper.}
		\label{fig:network_cat}
	\end{figure}

	Here we describe the different network architectures for integrating the target appearance, investigated in section \ref{network_study} of the main paper. Figure~\ref{fig:network_cat} visualizes the \textbf{Concatenation} architecture. In this architecture, both the reference and test branches have the same network structure. ResNet-18 \texttt{Block3} and  \texttt{Block4} features that are extracted from the reference and test images are passed through two \texttt{Conv} layers, followed by \texttt{PrPool} and an \texttt{FC} layer. The processed features from both the ResNet blocks and both the images are concatenated and passed through a final \texttt{FC} layer which predicts the IoU. Note that due to the symmetric structure of the network, the weights for the \texttt{Conv} layers before \texttt{PrPool} are shared between the reference branch and the test branch. However the \texttt{FC} layers do not share the weights.
	
\begin{figure}[t]
		\centering
		\includegraphics[width=\columnwidth,trim={0.0cm 0.0cm 0.0cm 0.0cm},clip]{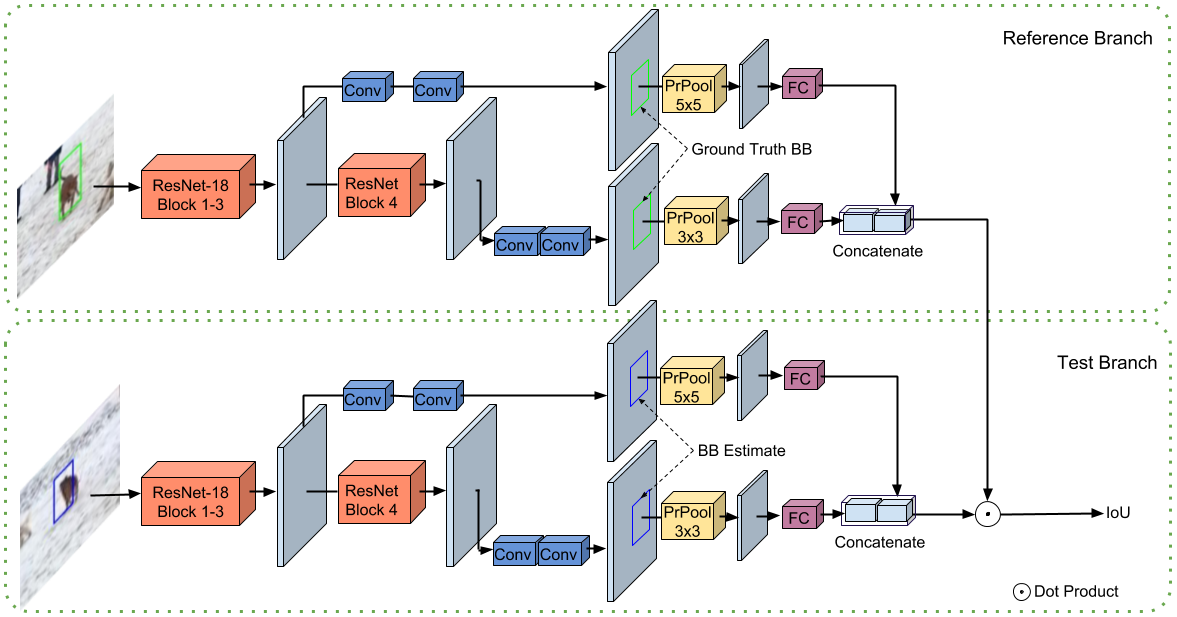}\vspace{-1mm}
		\vspace{-3mm}
		\caption{Architecture of the \textbf{Siamese} network for IoU prediction evaluated in section \ref{network_study} in the paper.}
		\label{fig:network_siam}
	\end{figure}

	Figure~\ref{fig:network_siam} visualizes the \textbf{Siamese} architecture. Similar to \textbf{Concatenation}, both the reference and test branches have the same network structure. ResNet-18 \texttt{Block3} and  \texttt{Block4} features that are extracted from the reference and test images are passed through two \texttt{Conv} layers, followed by \texttt{PrPool} and an \texttt{FC} layer. The processed features from both the ResNet blocks are then concatenated. The IoU prediction is obtained as the dot product of the features from the reference and the test branches. The \texttt{Conv} layers before \texttt{PrPool} have shared weights. In the final FC layer however, we found it beneficial not to share the weights between the branches. 
	
\section{Convergence Analysis}
\label{sec:convergence}

We empirically compare of the convergence speed of the employed optimization method (algorithm \ref{alg:GNCG} in the paper) and Gradient Descent (GD). This is performed by comparing the loss for the online learning problem  eq.\ \eqref{eq:classification-loss}, which is minimized in the first frame w.r.t.\ the filter weights $w_1$ and $w_2$. For our method, we use the settings described in the paper. In case of Gradient Descent we employ the same settings used in the ablation study (section \ref{method_ablation} in the paper).

\begin{figure}[t]
	\centering
	\includegraphics[width=\columnwidth,trim={0.0cm 0.0cm 0.0cm 0.0cm},clip]{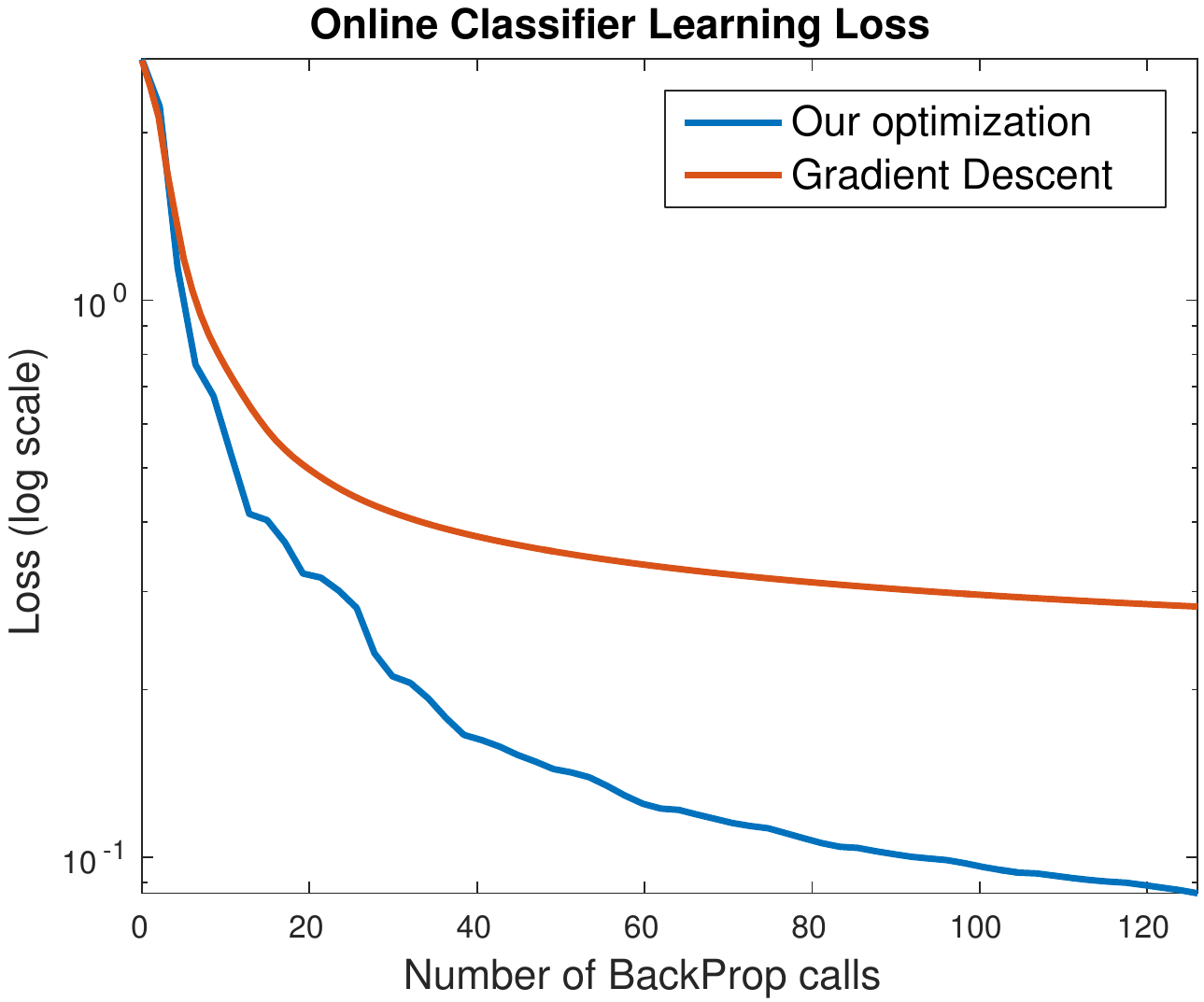}\vspace{-1mm}
	\vspace{-3mm}
	\caption{Comparison of convergence speed between our employed online optimization procedure and Gradient Descent. We plot the loss of the online classifier learning (eq.\ \eqref{eq:classification-loss} in the paper) w.r.t.\ the number of performed $\bp$ iterations. The loss is averaged over five independent runs of the complete NFS dataset. The employed method achieves much faster convergence.}
	\label{fig:convergence}
\end{figure}

In figure~\ref{fig:convergence} we plot the loss (eq.\ \eqref{eq:classification-loss} in the paper) for each method. For a fair comparison, the loss is plotted w.r.t.\ the number of $\bp$ calls performed by each method. The loss in figure~\ref{fig:convergence} is computed as an average of five complete runs over the full NFS dataset \cite{NfS}. Our CG-based optimization algorithm exhibits superior convergence speed compared to Gradient Descent. Moreover, the employed optimization methods does not require tuning of the step length and momentum parameters.

\section{Detailed results on LaSOT dataset}

\begin{figure}[t]
	\centering%
	\newcommand{\wid}{0.9\columnwidth}%
	\includegraphics*[trim = 20 120 20 130, width = \wid]{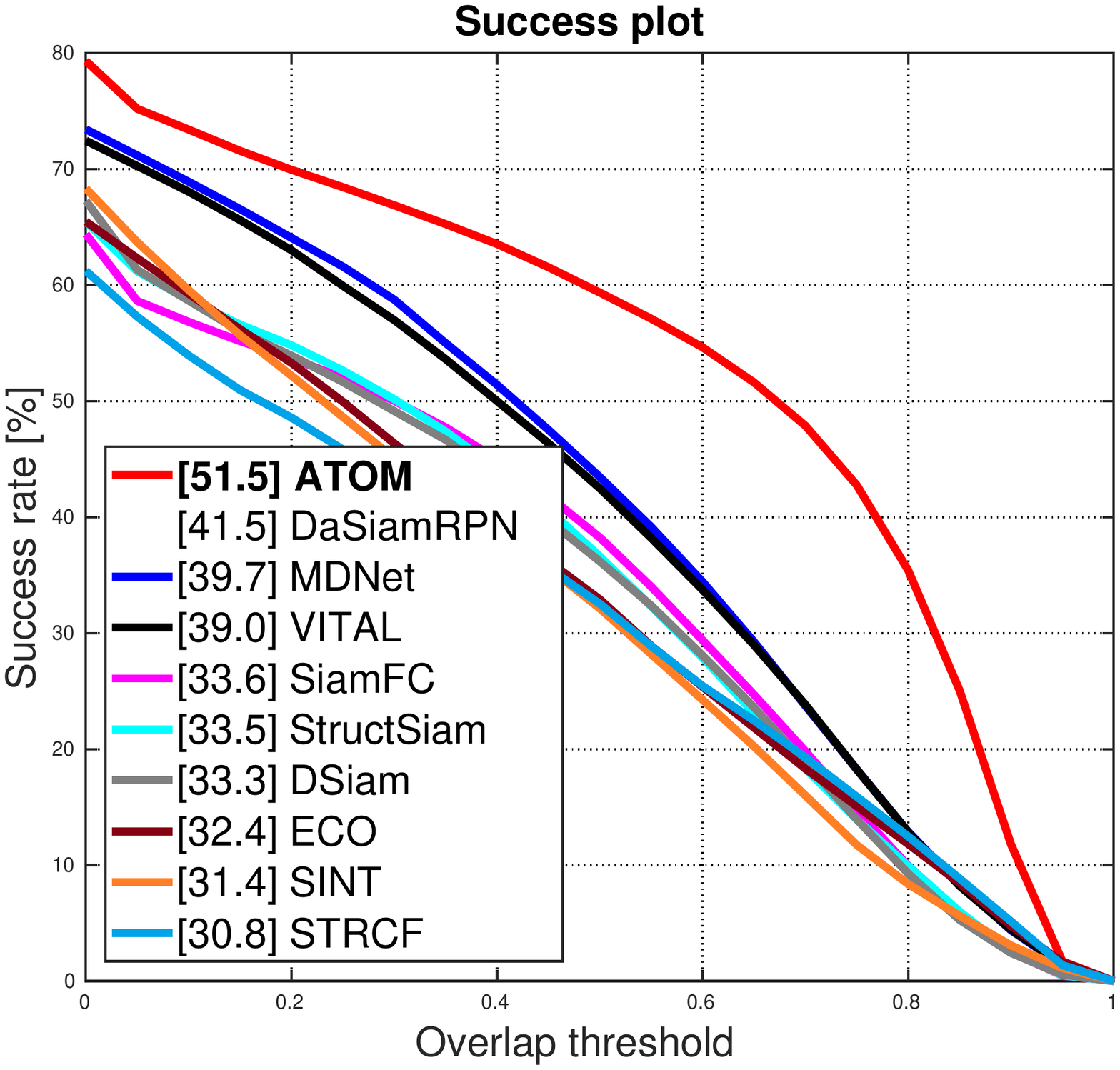}\vspace{-1.5mm}%
	\caption{Success plot on the LaSOT dataset. Note that due to the unavailability of raw results for DaSiamRPN, we only report the final AUC score in the legend. Our approach ATOM outperforms all previous methods by a large margin.
	}%
	\label{fig:lasot_success}\vspace{-1mm}%
\end{figure}

\label{sec:lasot}
In table \ref{tab:lasot} in the main paper, we provide a state-of-the-art comparison on the large-scale LaSOT dataset in terms of normalized precision and success. Here, we provide the success plot for the same. The success plots are obtained using the overlap precision (OP) score, which is computed as the percentage of frames in the dataset for which the intersection-over-union (IoU) overlap between the tracker prediction and the ground truth bounding box is higher than a certain threshold. The OP scores are plotted for a range of thresholds in $[0,1]$ to obtain the success plot. The area under this plot gives the AUC (success) score, which is reported in the legend. Figure \ref{fig:lasot_success} shows the success plot over the 280 test videos. Our approach ATOM significantly outperforms the previous best approach DaSiamRPN~\cite{DaSiamRPN} with an absolute gain of $10.0\%$ in AUC score.

\section{Results on OTB-100 dataset}
\label{sec:otb}

\begin{figure}[t]
	\centering
	\includegraphics[width=0.9\columnwidth,trim={0.0cm 0.0cm 0.0cm 0.0cm},clip]{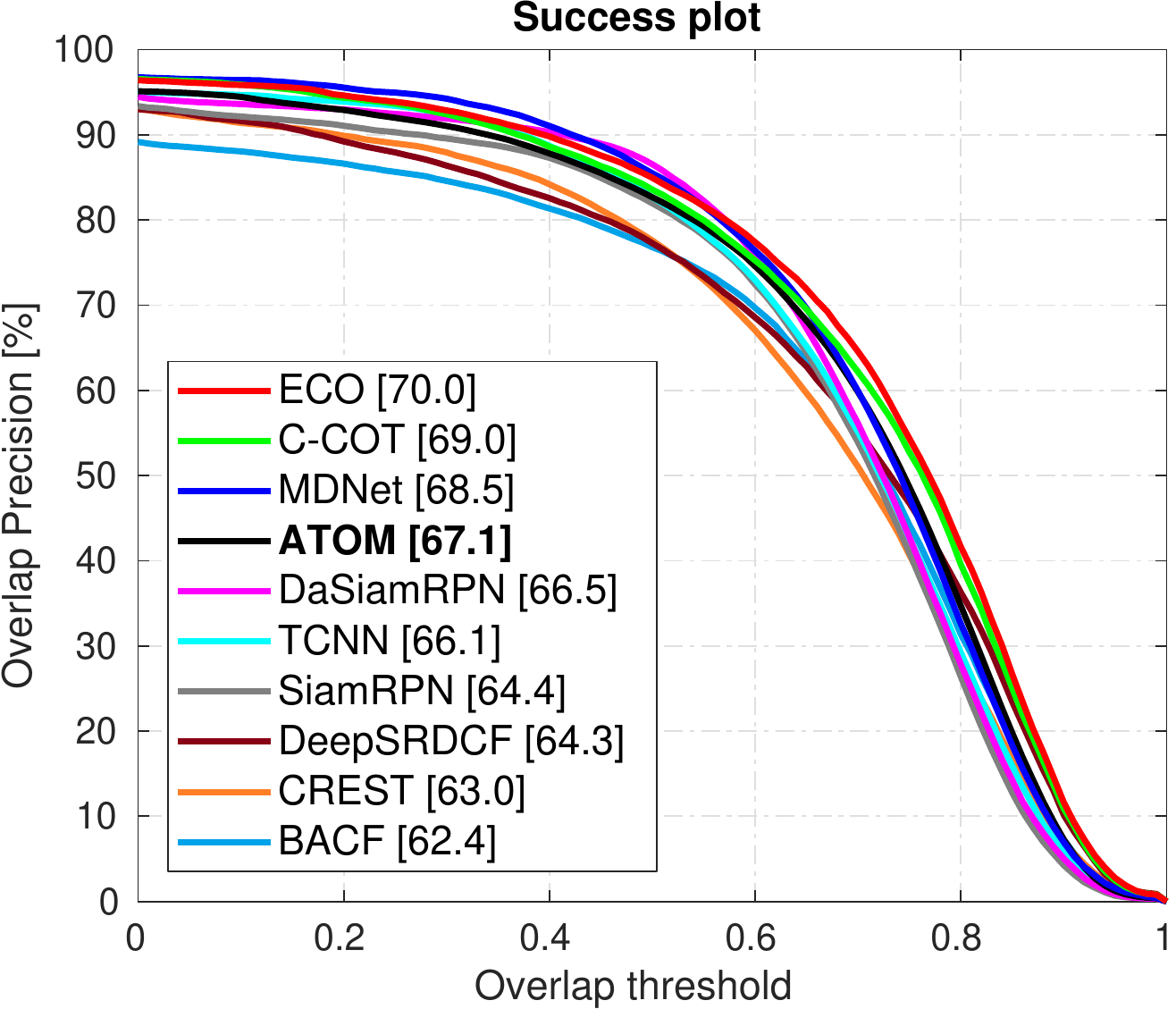}
	\vspace{0mm}
	\caption{State-of-the-art comparison on the OTB-100 dataset. Our approach obtains results competitive with the state-of-the-art approaches.}
	\label{fig:otb}
\end{figure}

Here, we compare our approach with the state-of-the-art trackers on the OTB-100 \cite{OTB2015} dataset. The success plot over all the 100 videos are shown in figure \ref{fig:otb}. Our approach achieves results competitive with the state-of-the-art approaches, with an AUC score of $67.1\%$. Note that the best results are obtained by the correlation filter based methods, ECO~\cite{DanelljanCVPR2017} and CCOT~\cite{DanelljanECCV2016}. These methods employ brute-force multi-scale search for target estimation. Since OTB-100 has limited changes in aspect ratio (see figure 2 in~\cite{UAV123}), the fixed aspect ratio constraint in multi-scale search strategy helps these methods to obtain a better accuracy.

\section{Impact of training data}
\label{sec:training_data}

\begin{table}[!t]
	\centering\vspace{-1mm}
	\resizebox{1.01\columnwidth}{!}{%
		\begin{tabular}{l@{~}c@{~~}c@{~~}c@{~~}c@{~~}c@{~~}c@{~~}c@{~~}c@{~~}c@{~~}c@{~~}}
\toprule
&SINT&ECO&DSiam&StructSiam&SiamFC&VITAL&MDNet&DaSiamRPN&\textbf{ATOM-VID}&\textbf{ATOM}\\\midrule
Norm.\ Prec. (\%)&35.4&33.8&40.5&41.8&42.0&45.3&46.0&49.6&\textbf{\textcolor{blue}{55.0}}&\textbf{\textcolor{red}{57.6}}\\
Success (\%)&31.4&32.4&33.3&33.5&33.6&39.0&39.7&41.5&\textbf{\textcolor{blue}{49.5}}&\textbf{\textcolor{red}{51.5}}\\\bottomrule
\end{tabular}

	}\vspace{1mm}%
	\caption{Comparision of our approach trained using only ImageNet-VID (denoted ATOM-VID) on the LaSOT dataset. Our approach, trained using considerably less data as compared to the previous best approach DaSiamRPN, significantly outperforms it with an absolute gain of $8.0\%$ in AUC score.}
	\label{tab:lasot_vid}%
	\vspace{-1mm}
\end{table}

\begin{table}[!t]
	\centering\vspace{-1mm}
	\resizebox{1.01\columnwidth}{!}{%
		\begin{tabular}{l@{~}c@{~~}c@{~~}c@{~~}c@{~~}c@{~~}c@{~~}c@{~~}c@{~~}c@{~~}c@{~~}c@{~~}}
\toprule
&Staple&SAMF&CSRDCF&ECO&SiamFC&CFNet&MDNet&UPDT&DaSiamRPN&\textbf{ATOM-VID}&\textbf{ATOM}\\\midrule
Precision (\%)&47.0&47.7&48.0&49.2&53.3&53.3&56.5&55.7&59.1&\textbf{\textcolor{blue}{61.8}}&\textbf{\textcolor{red}{64.8}}\\
Norm.\ Prec. (\%)&60.3&59.8&62.2&61.8&66.6&65.4&70.5&70.2&73.3&\textbf{\textcolor{blue}{74.6}}&\textbf{\textcolor{red}{77.1}}\\
Success (\%)&52.8&50.4&53.4&55.4&57.1&57.8&60.6&61.1&63.8&\textbf{\textcolor{blue}{69.8}}&\textbf{\textcolor{red}{70.3}}\\\bottomrule
\end{tabular}

	}\vspace{1mm}%
	\caption{Comparision of our approach trained using only ImageNet-VID (denoted ATOM-VID) on the TrackingNet dataset.}
	\label{tab:tn}%
	\vspace{-1mm}
\end{table}

In this section, we investigate the impact of using recent large-scale tracking datasets for offline training of our IoU predictor network. We train our network using only the ImageNet-VID \cite{ILSVRC15} dataset, that has been commonly used to train trackers \cite{SiameseFC,Valmadre2017cvpr,RTINet} in recent years. We compare this version, denoted ATOM-VID, with the state-of-the-art approaches on two recent datasets, namely LaSOT \cite{LaSOT} and TrackingNet \cite{TrackingNet}. For comparision, we also include our final version ATOM, trained using the train splits of LaSOT, TrackingNet and COCO \cite{COCO}. Results are shown in table \ref{tab:lasot_vid} for LaSOT and table \ref{tab:tn} for TrackingNet, respectively. Among previous approaches, DaSiamRPN \cite{DaSiamRPN} uses bounding box regression strategy and achieves the best results on both datasets. Note that DaSiamRPN is trained using the large-scale YoutubeBB \cite{YoutubeBB}, ImageNet-VID, COCO and ImageNet DET \cite{ILSVRC15} datasets. Our approach ATOM-VID, trained using only ImageNet-VID, significantly outperforms DaSiamRPN with an absolute gain of $8.0\%$ in AUC score on LaSOT, and $6.0\%$ in AUC score on TrackingNet. Using the recent tracking datasets for training further improves the results, providing an absolute gain of $2.0\%$ on LaSOT and $0.5\%$ on TrackingNet. While using a larger training set improves the tracking performance as expected, our approach still achieves state-of-the-art results when using less data compared to recent methods.

\section{Additional Results on UAV123}
\label{sec:additional_results}
	
\begin{figure*}[!t]
	\centering
	\newcommand{\wid}{0.33\textwidth}
	\newcommand{\name}{figures/att_uav}
	\newcommand{\eval}{OPE}
	\includegraphics[width=\wid]{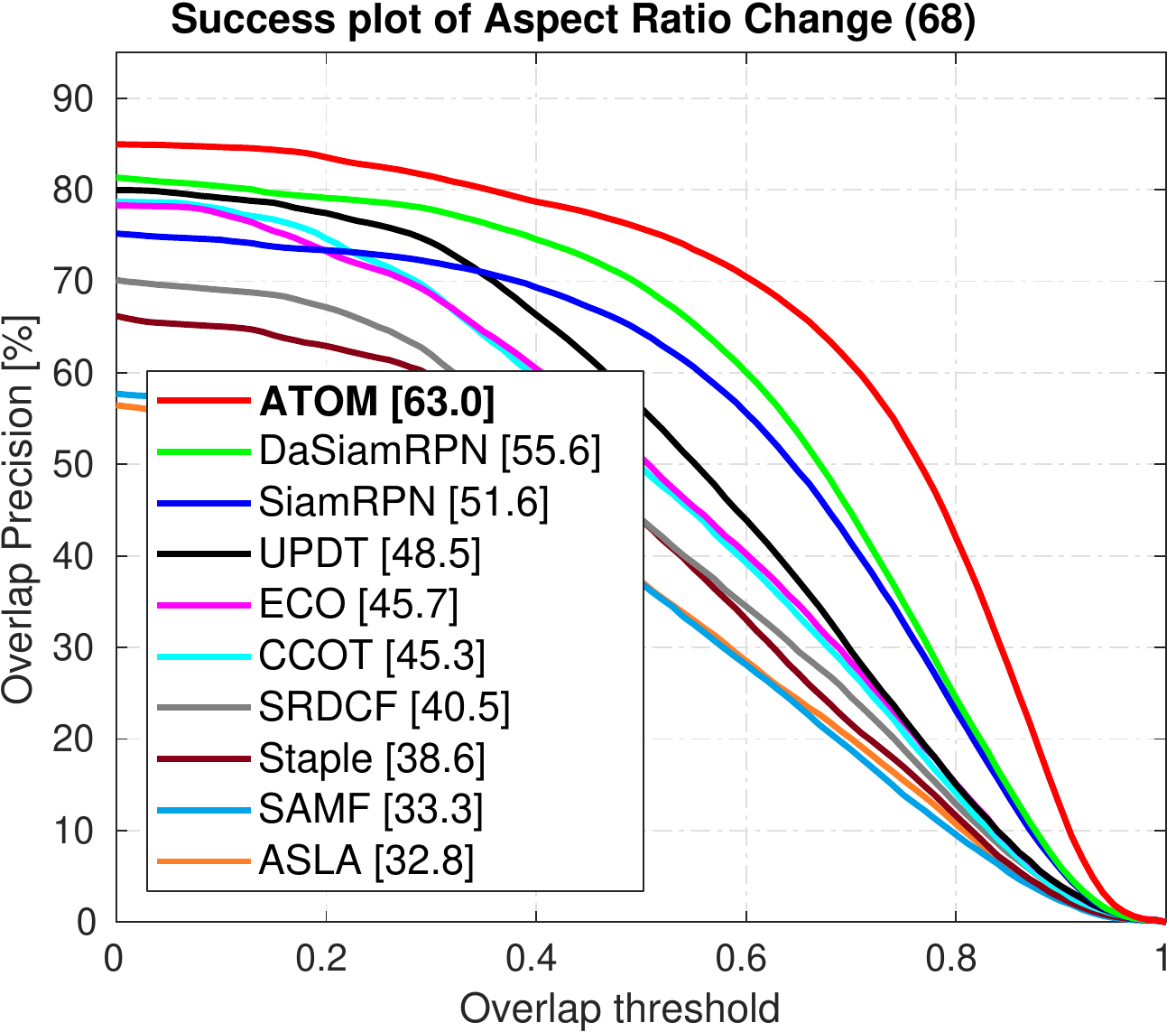}%
	\includegraphics[width=\wid]{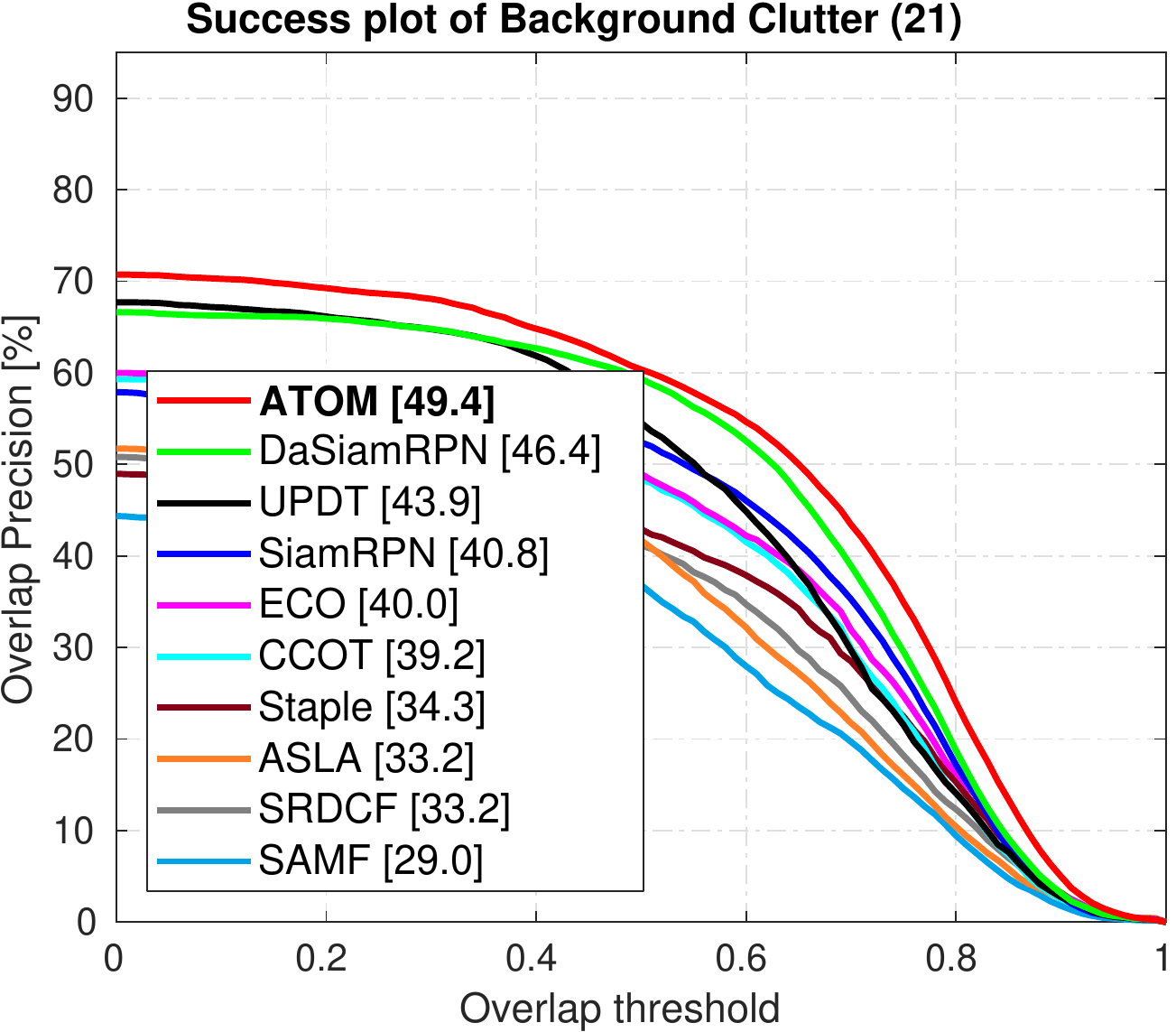}%
	\includegraphics[width=\wid]{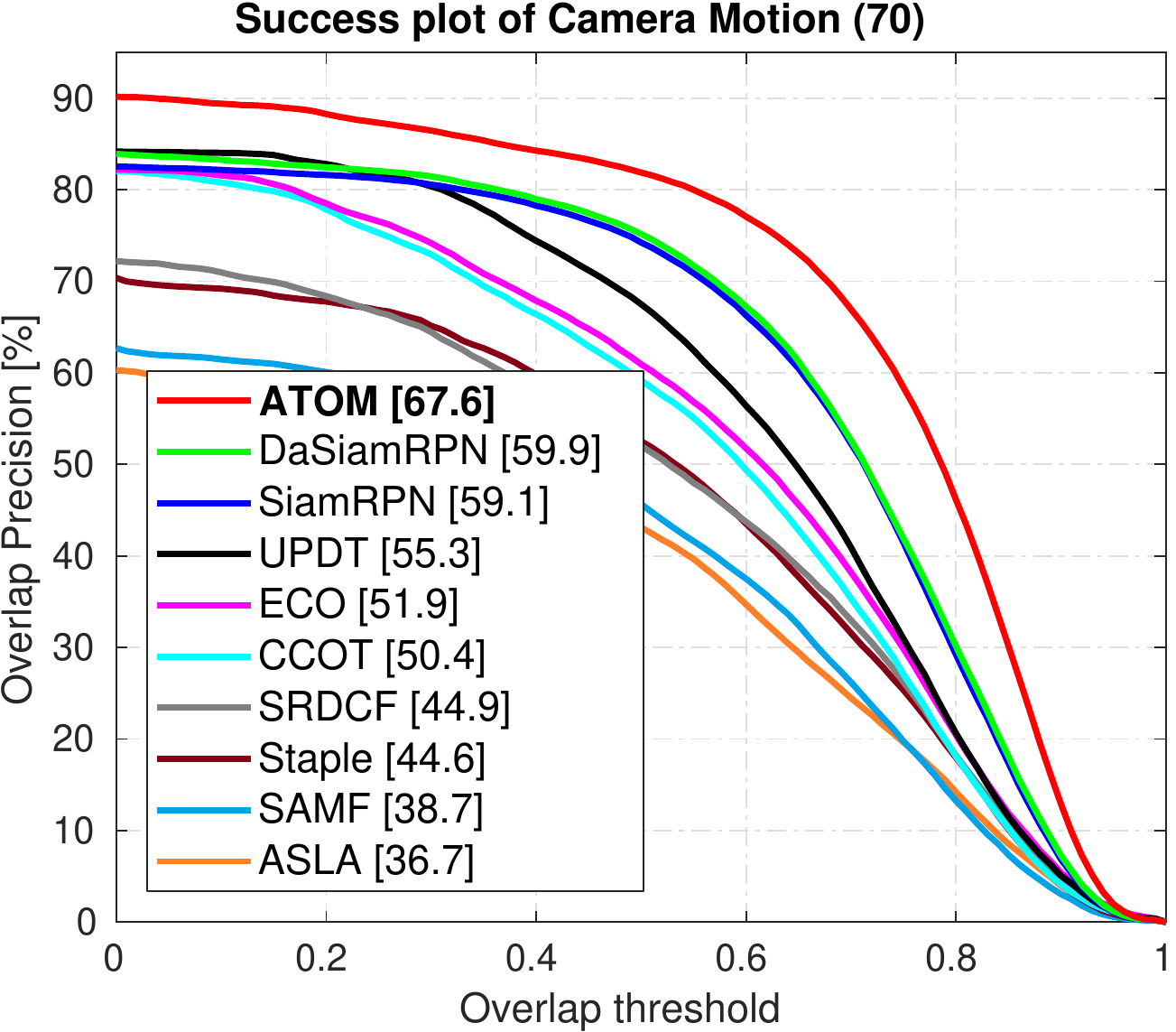}
	\includegraphics[width=\wid]{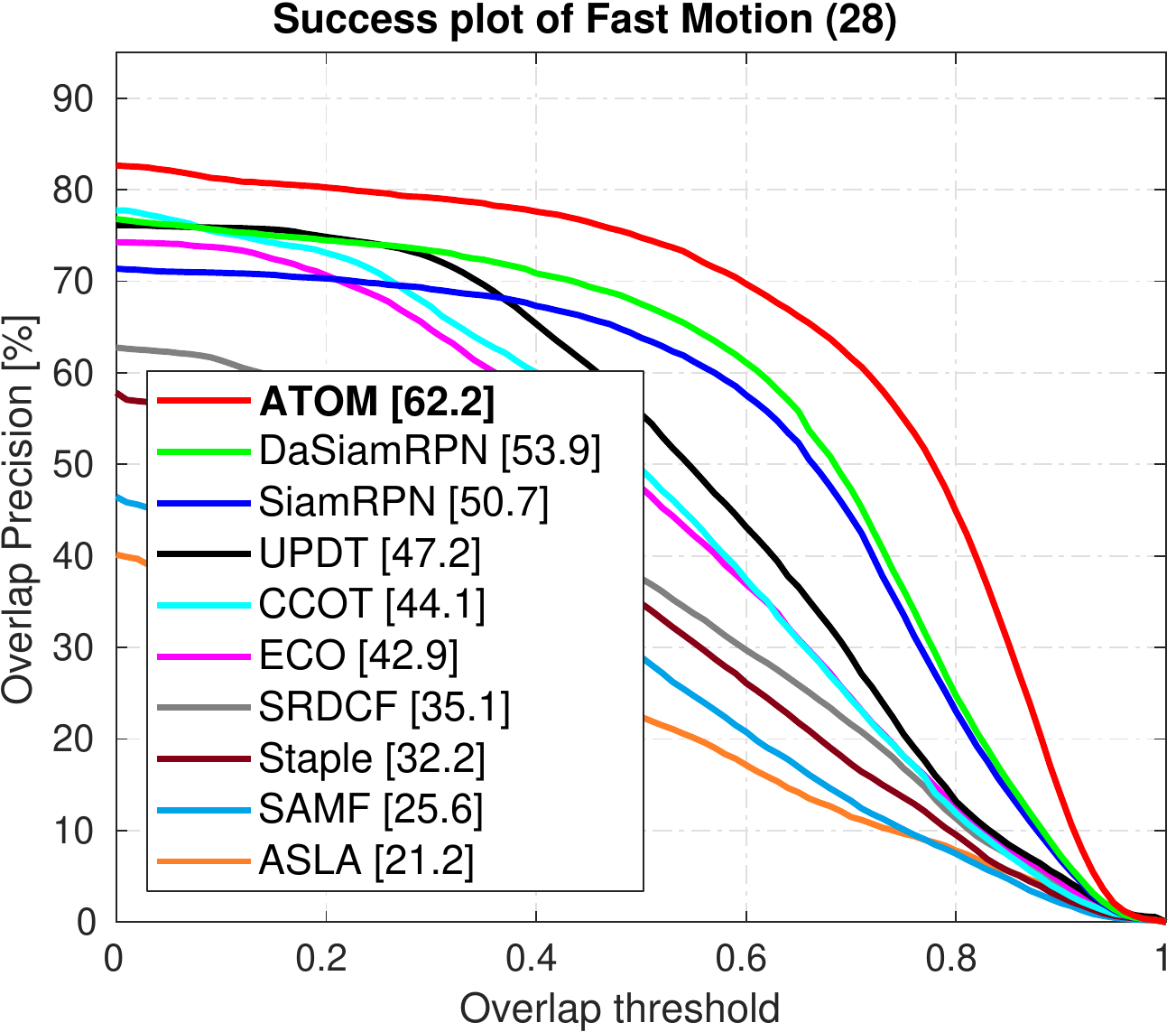}%
	\includegraphics[width=\wid]{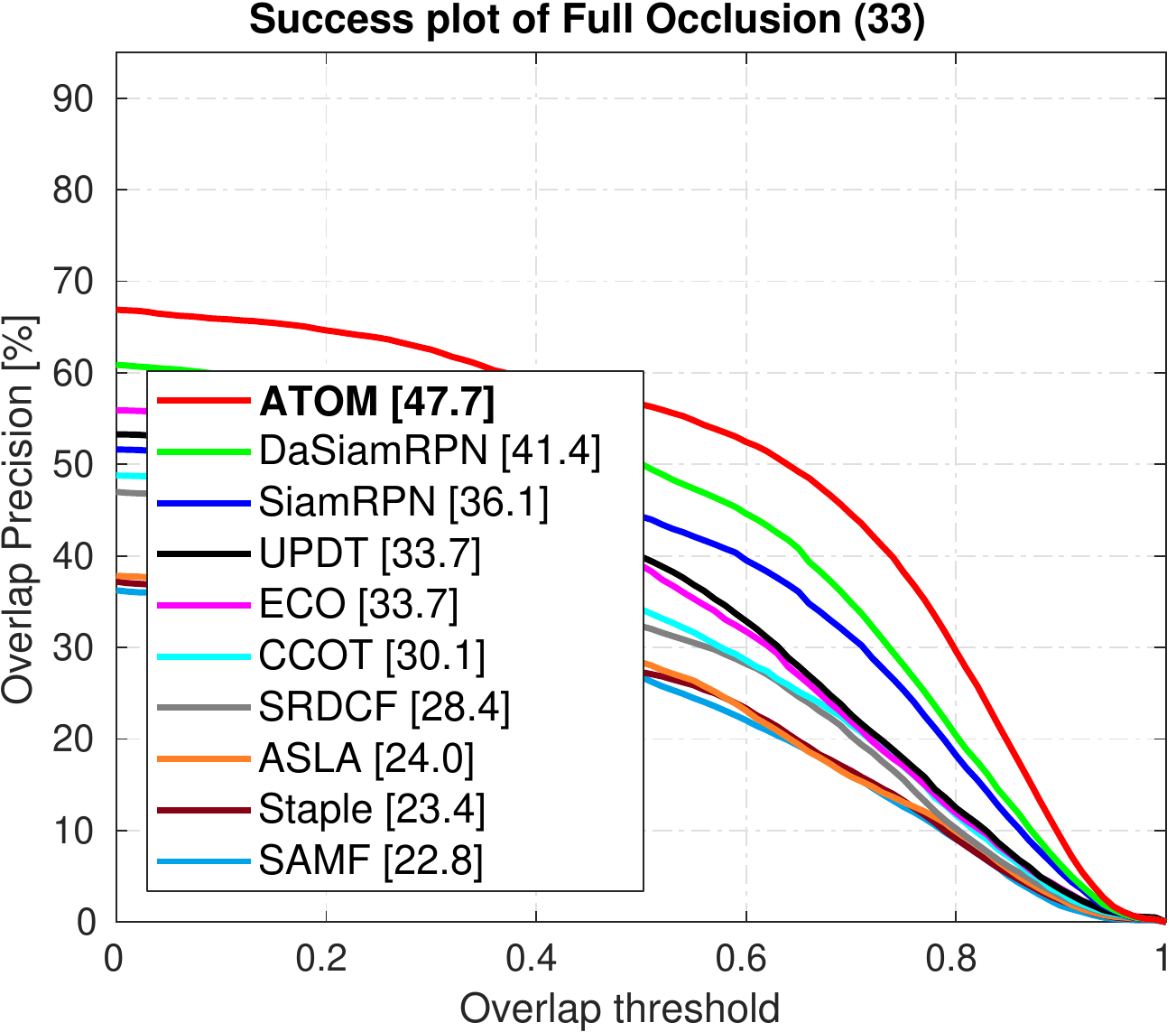}%
	\includegraphics[width=\wid]{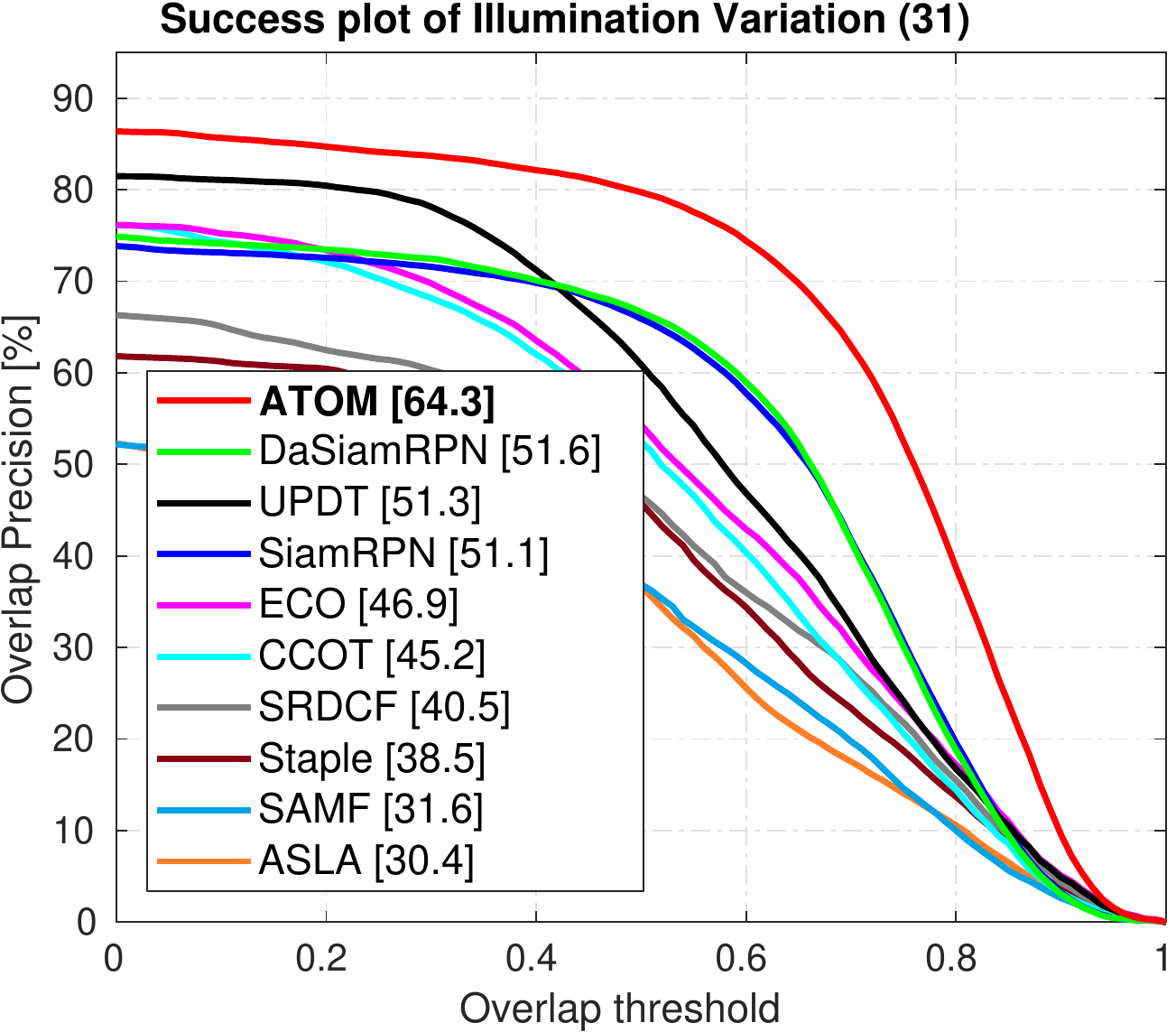}
	\includegraphics[width=\wid]{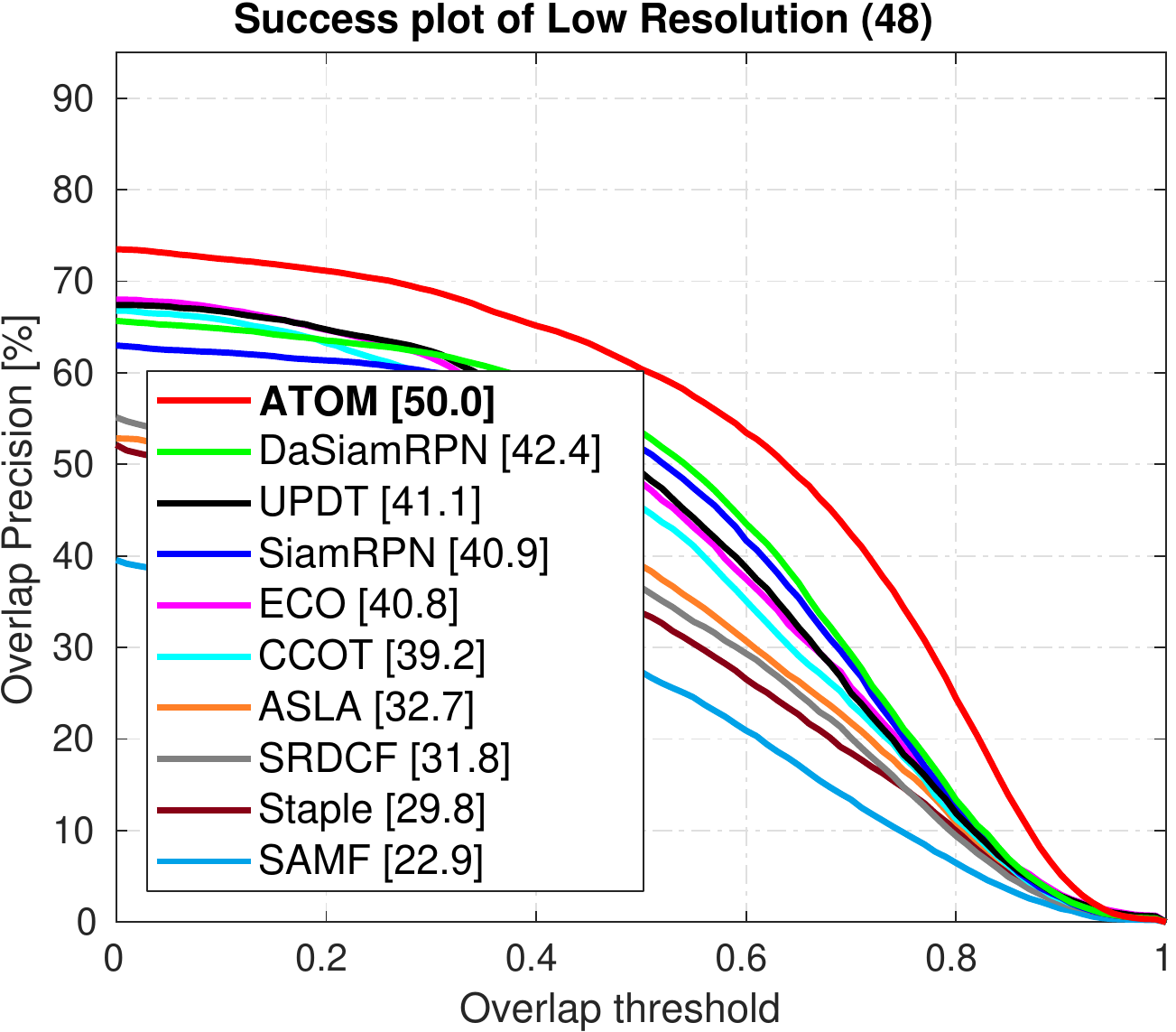}%
	\includegraphics[width=\wid]{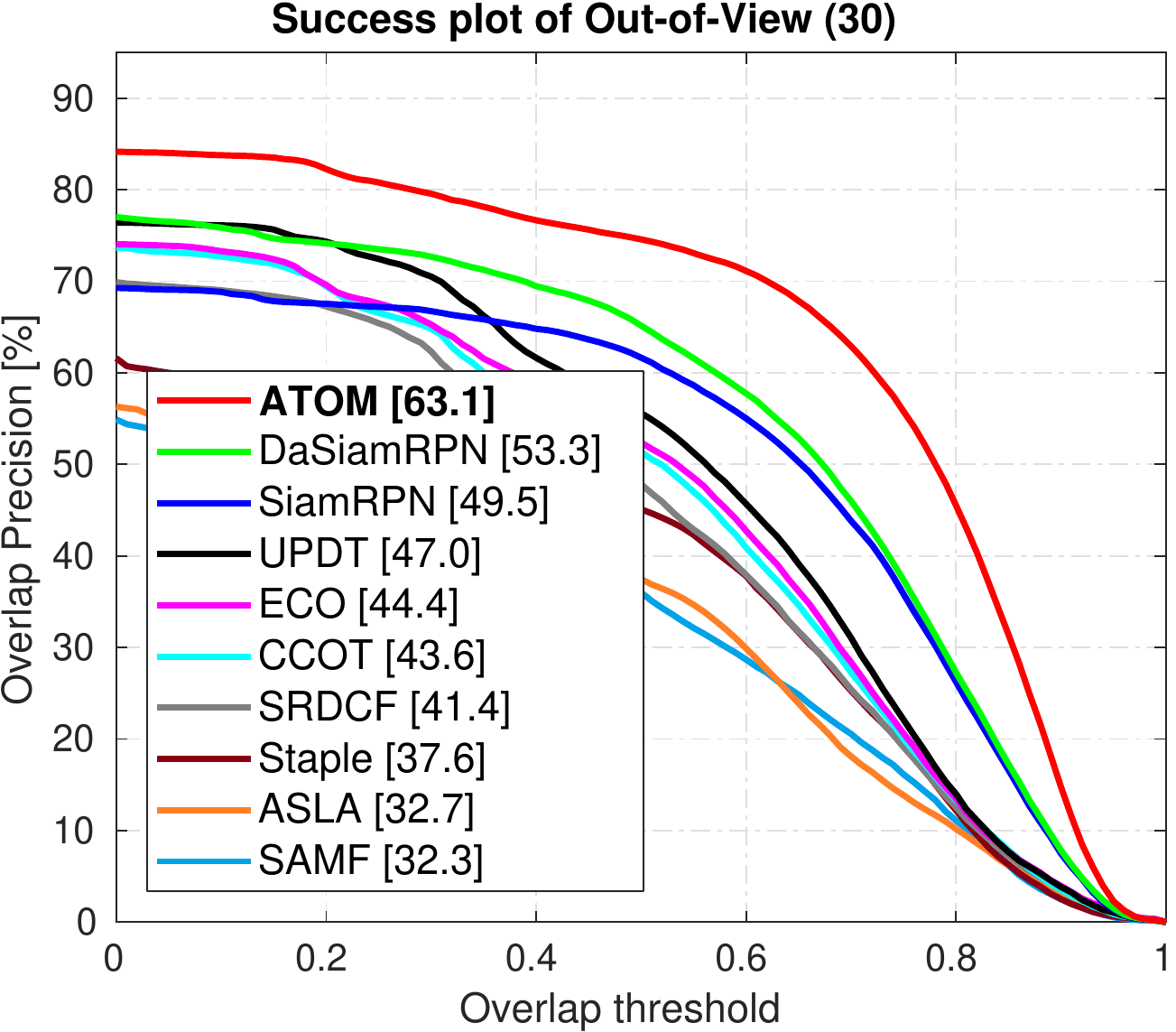}%
	\includegraphics[width=\wid]{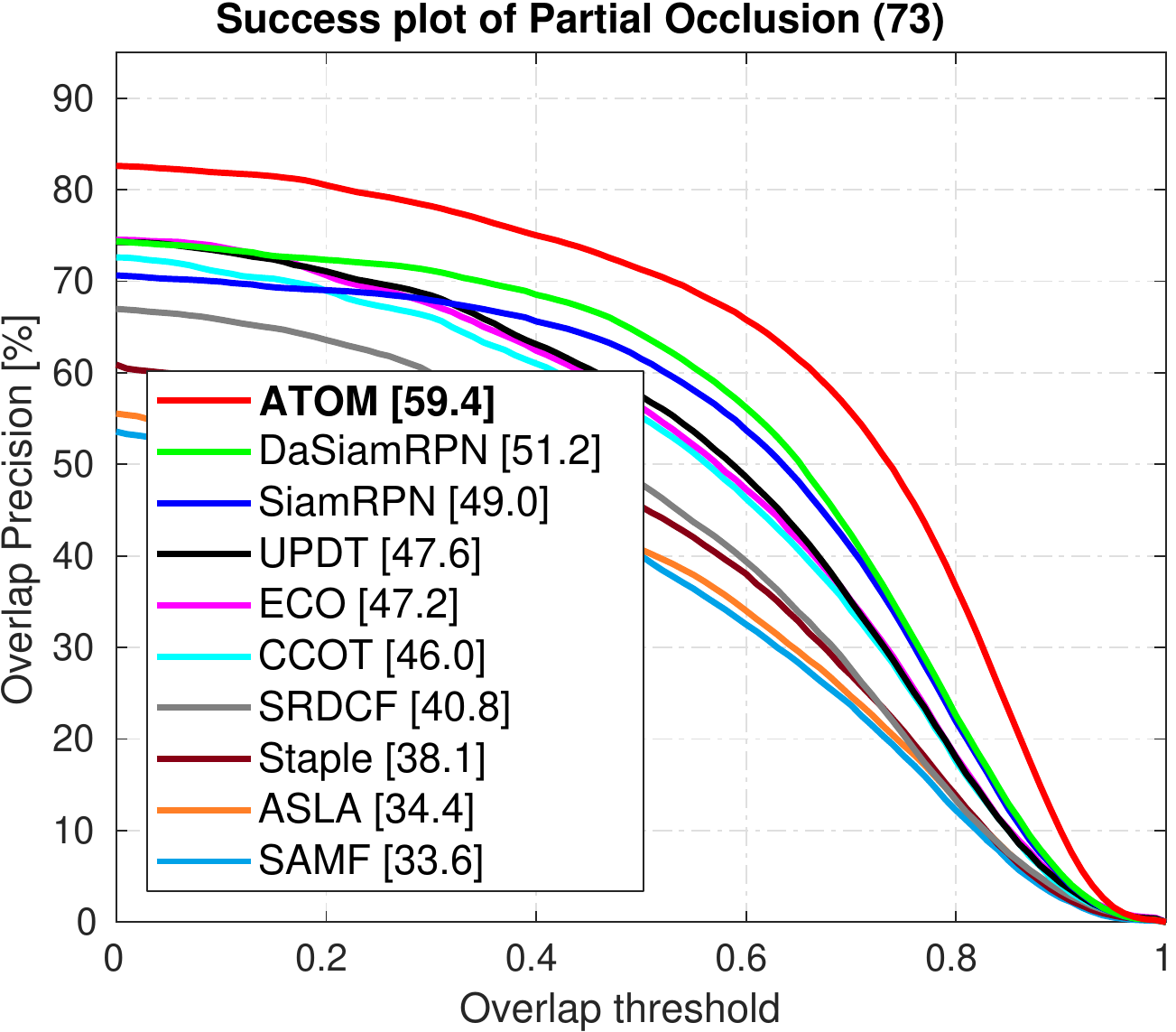}
	\includegraphics[width=\wid]{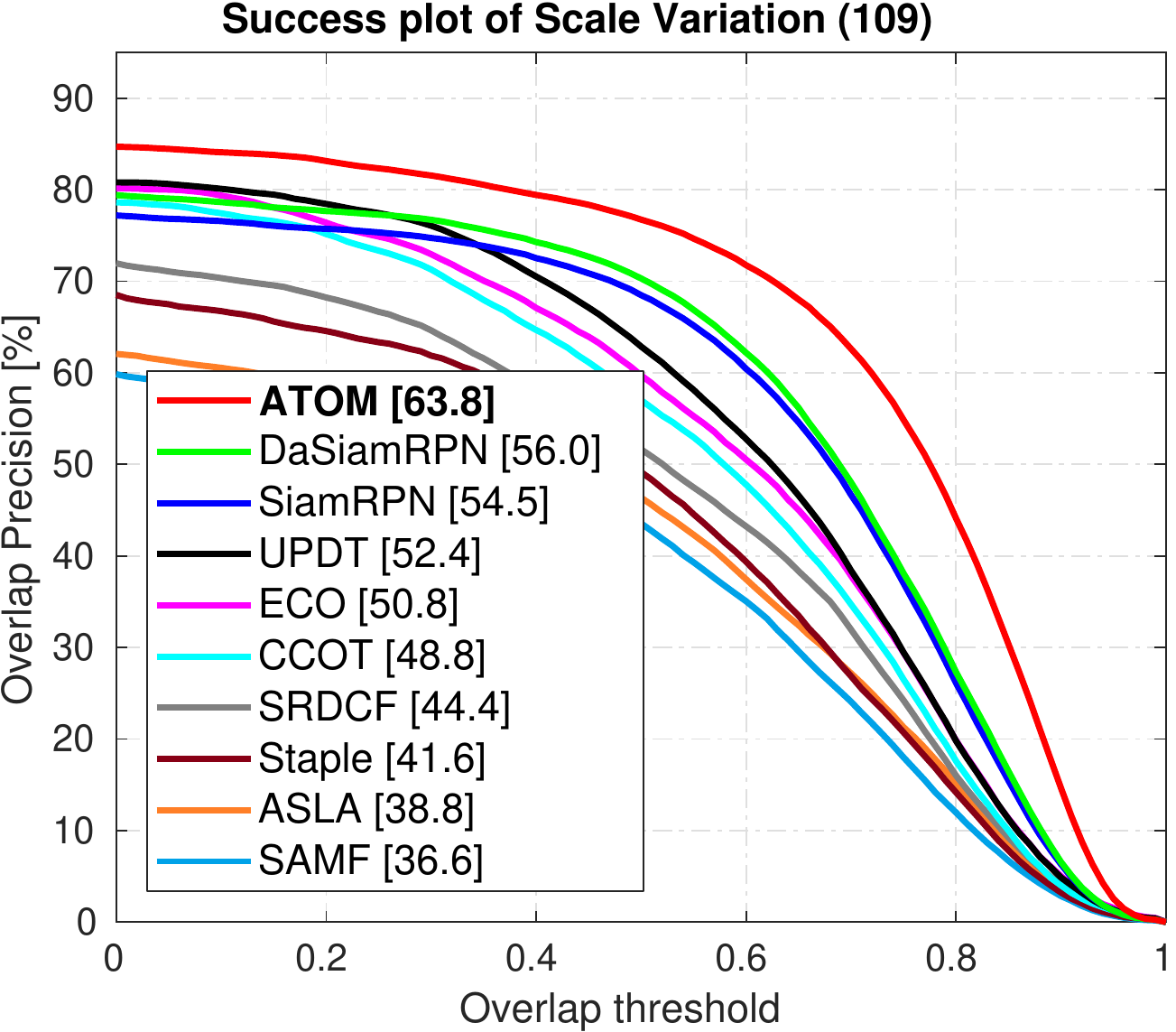}%
	\includegraphics[width=\wid]{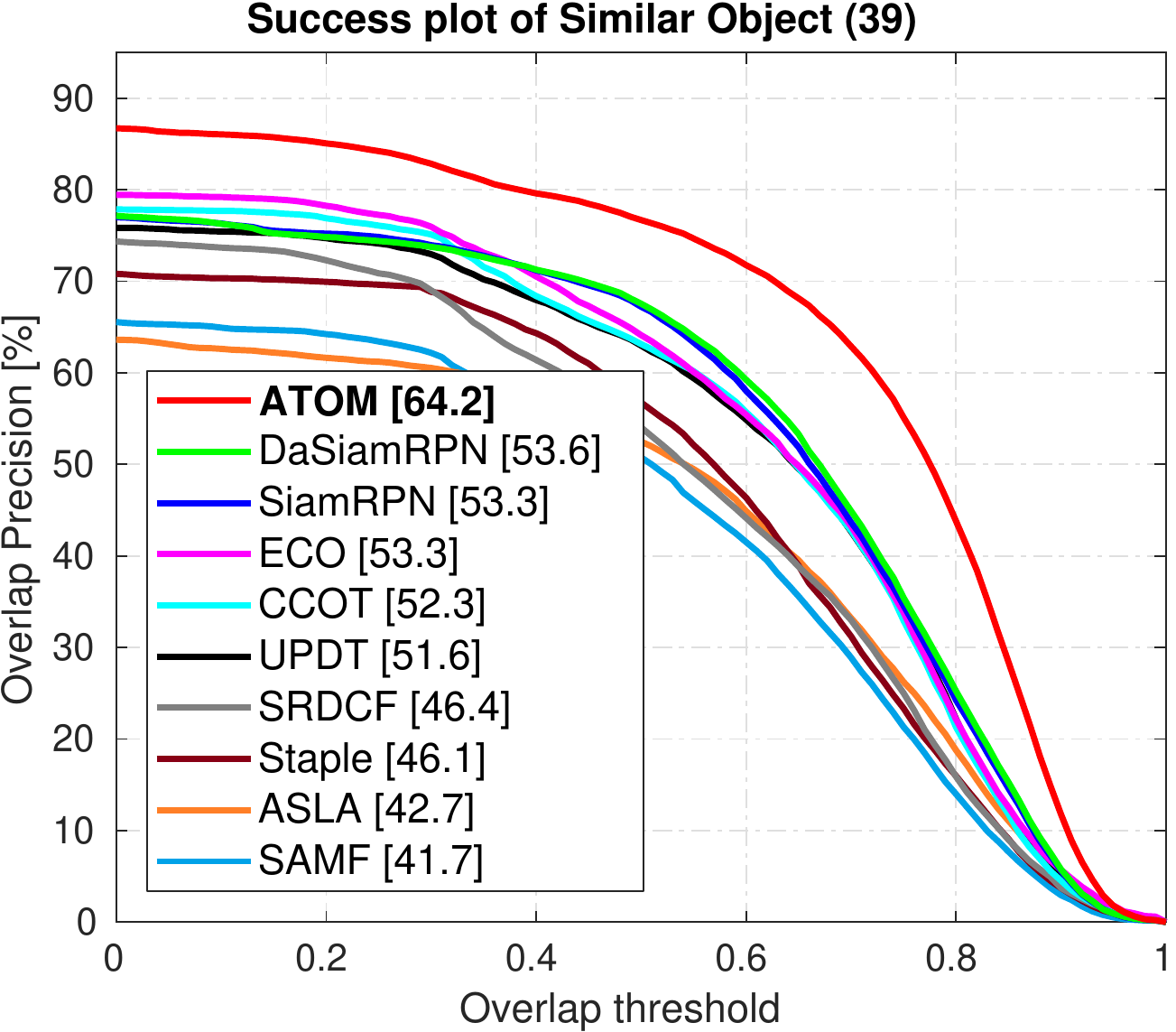}%
	\includegraphics[width=\wid]{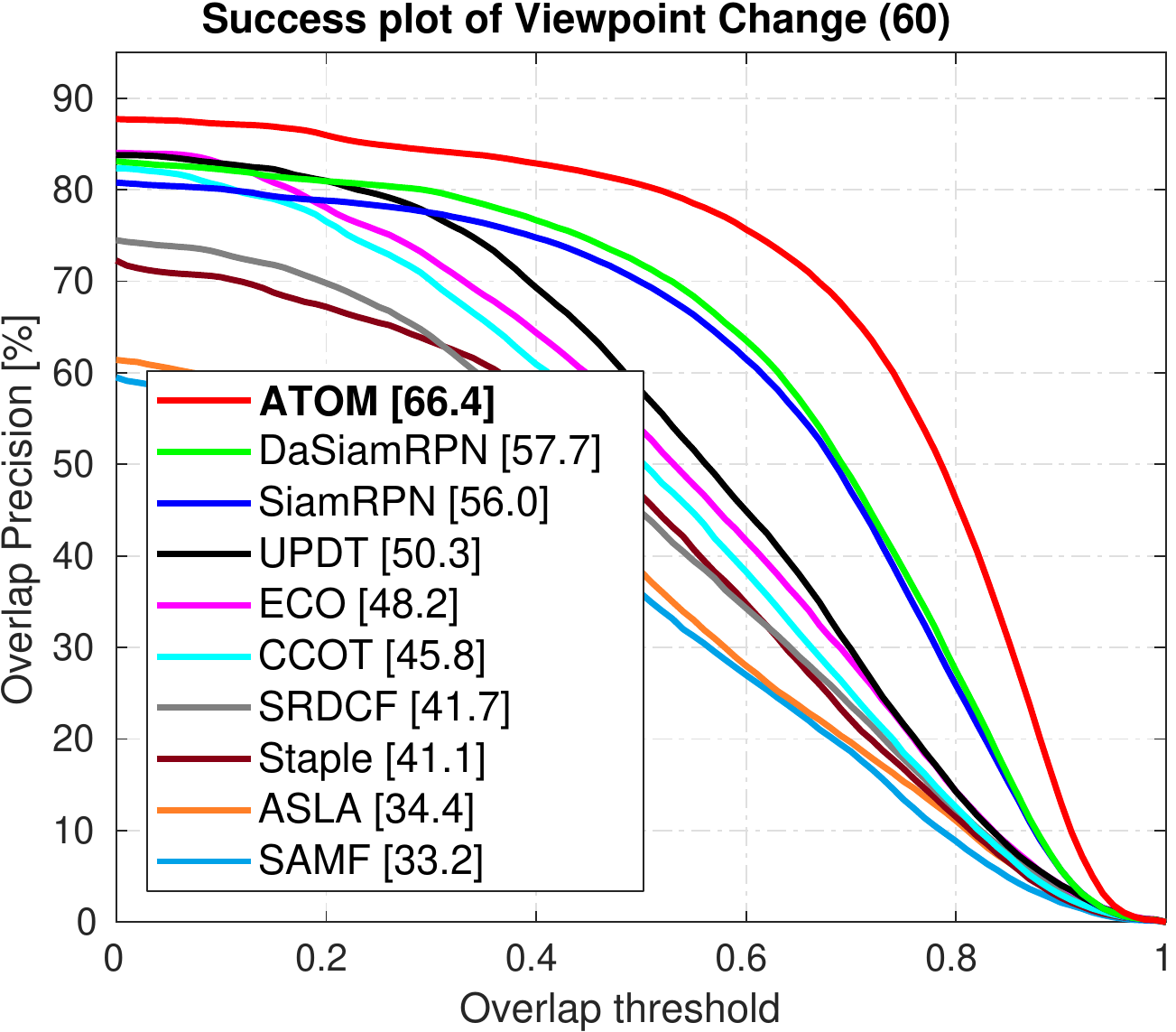}
	\caption{Attribute analysis on the UAV123 dataset. Our approach \textbf{ATOM} obtains the best performance on all 12 attributes.}
	\label{fig:attribute_uav}
\end{figure*}
	
	Here, we provide detailed results on the UAV123 dataset~\cite{UAV123}. In UAV123, each video is annotated with $12$ different attributes: aspect ratio change, background clutter, camera motion, fast motion, full occlusion, illumination variation, low resolution, out-of-view, partial occlusion, scale variation, similar objects, and viewpoint change. Figure \ref{fig:attribute_uav} shows the success plots for all the attributes. Our approach obtains the best results on all $12$ attributes. Thanks to our target estimation module, our approach excels in case of aspect ratio change, scale variation, and viewpoint change. Furthermore, due to our robust online-learned classifier, our tracker also outperforms previous methods in case of similar objects, illumination variation, partial occlusion, and low resolution.

\end{document}